%% file: main.tex
\documentclass[runningheads]{llncs}
\usepackage{graphicx}
\usepackage{amsmath}
\usepackage{amssymb}
\usepackage{booktabs}
\usepackage{multirow}
\usepackage{makecell}
\usepackage{algorithm}
\usepackage{algorithmic}
\usepackage[table]{xcolor}
\definecolor{Gray}{gray}{.9}
\definecolor{White}{gray}{1.}
\newcolumntype{g}{>{\columncolor{Gray}}c}
\usepackage{caption}
\usepackage{subcaption}
\usepackage{tikz}
\usepackage{comment}
\usepackage{color}

\usepackage[pagebackref,breaklinks,colorlinks]{hyperref}

\usepackage[capitalize]{cleveref}
\crefname{section}{Sec.}{Secs.}
\Crefname{section}{Section}{Sections}
\Crefname{table}{Table}{Tables}
\crefname{table}{Tab.}{Tabs.}

\usepackage[accsupp]{axessibility}  

\def\printappendix{}
\ifdefined\printappendix
    
    \def\appendixa{\Cref{sec:natural_k_shot}}
    \def\tablethree{\Cref{tab:5_shot}}
\else
    
    \def\appendixa{the supplementary material}
    \def\tablethree{Table~\textcolor{red}{3}}
\fi

\usepackage{xspace}

\makeatletter
\DeclareRobustCommand\onedot{\futurelet\@let@token\@onedot}
\def\@onedot{\ifx\@let@token.\else.\null\fi\xspace}

\def\eg{\emph{e.g}\onedot} 
\def\ie{\emph{i.e}\onedot} 
 
\def\etc{\emph{etc}\onedot} 
 
\def\etal{\emph{et al}\onedot}
\makeatother

\begin{document}
\pagestyle{headings}
\mainmatter

\newcommand\blfootnote[1]{%
  \begingroup
  \renewcommand\thefootnote{}\footnote{#1}%
  \addtocounter{footnote}{-1}%
  \endgroup
}

\title{Rethinking Few-Shot Object Detection\texorpdfstring{\\}{ }on a Multi-Domain Benchmark}
\titlerunning{Rethinking Few-Shot Object Detection on a Multi-Domain Benchmark}
\author{%
Kibok Lee\inst{1, 2}$^\star$ \and%
Hao Yang\inst{1}$^\dagger$ \and%
Satyaki Chakraborty\inst{1} \and%
Zhaowei Cai\inst{1} \and\\%
Gurumurthy Swaminathan\inst{1} \and%
Avinash Ravichandran\inst{1} \and%
Onkar Dabeer\inst{1}%
}
\authorrunning{K. Lee et al.}
\institute{%
$^1$AWS AI Labs \quad $^2$Yonsei University \\
\email{\{kibok,haoyng,satyaki,zhaoweic,gurumurs,ravinash,onkardab\}@amazon.com} \\
\email{kibok@yonsei.ac.kr}
}
\maketitle

\begin{abstract}
Most existing works on few-shot object detection (FSOD) focus on a setting where both pre-training and few-shot learning datasets are from a similar domain. However, few-shot algorithms are important in multiple domains; hence evaluation needs to reflect the broad applications. We propose a Multi-dOmain Few-Shot Object Detection (MoFSOD) benchmark consisting of 10 datasets from a wide range of domains to evaluate FSOD algorithms. We comprehensively analyze the impacts of freezing layers, different architectures, and different pre-training datasets on FSOD performance. Our empirical results show several key factors that have not been explored in previous works: 1) contrary to previous belief, on a multi-domain benchmark, fine-tuning (FT) is a strong baseline for FSOD, performing on par or better than the state-of-the-art (SOTA) algorithms; 2) utilizing FT as the baseline allows us to explore multiple architectures, and we found them to have a significant impact on down-stream few-shot tasks, even with similar pre-training performances; 3) by decoupling pre-training and few-shot learning, MoFSOD allows us to explore the impact of different pre-training datasets, and the right choice can boost the performance of the down-stream tasks significantly. Based on these findings, we list possible avenues of investigation for improving FSOD performance and propose two simple modifications to existing algorithms that lead to SOTA performance on the MoFSOD benchmark.
The code is available \href{https://github.com/amazon-research/few-shot-object-detection-benchmark}{here}.

\end{abstract}
\blfootnote{$^\star$Work done at AWS. $^\dagger$Corresponding author.}

\input{1_intro}

\input{2_related}

\input{3_fsod}

\input{4_exp}

\input{5_conclusion}


\clearpage
{\small
\bibliographystyle{splncs04}
\bibliography{ref}
}

\clearpage
\input{appendix}

\end{document}

%% file: 1_intro.tex
\section{Introduction}
\label{sec:intro}

Convolutional neural networks have led to significant progress in object detection by learning with a large number of training images with annotations~\cite{ren2015faster,redmon2017yolo9000,cai2018cascade,carion2020end}. However, humans can easily localize and recognize new objects with only a few examples. Few-shot object detection (FSOD) is a task to address this setting~\cite{kang2019few,wang2020frustratingly,fan2020fsod,sun2021fsce,qiao2021defrcn}. FSOD is desirable for many real-world applications in diverse domains due to lack of training data, difficulties in annotating them, or both, \eg, identifying new logos, detecting anomalies in the manufacturing process or rare animals in the wild, \etc. These diverse tasks naturally have vast differences in class distribution and style of images. Moreover, large-scale pre-training datasets in the same domain are not available for many of these tasks. In such cases, we can only rely on existing natural image datasets, such as COCO~\cite{COCO} and OpenImages~\cite{OpenImages} for pre-training.

Despite the diverse nature of FSOD tasks, FSOD benchmarks used in prior works are limited to a homogeneous setting~\cite{kang2019few,wang2020frustratingly,fan2020fsod,sun2021fsce,qiao2021defrcn}, such that the pre-training and few-shot test sets in these benchmarks are from the same domain, or even the same dataset, \eg, VOC~\cite{VOC} $15+5$ and COCO~\cite{COCO} $60+20$ splits. The class distributions of such few-shot test sets are also fixed to be balanced.
While they provide an artificially balanced environment for evaluating different algorithms, 
it might lead to skewed conclusions for applying them in more realistic scenarios. Note that few-shot classification suffered from the same problem in the past few years~\cite{vinyals16mini,ren2018tiered};
Meta-dataset~\cite{triantafillou2020meta} addressed the problem with 10 different domains and a sophisticated scheme to sample imbalanced few-shot episodes. 

\begin{figure}[t]
\centering
\includegraphics[width=0.80\linewidth]{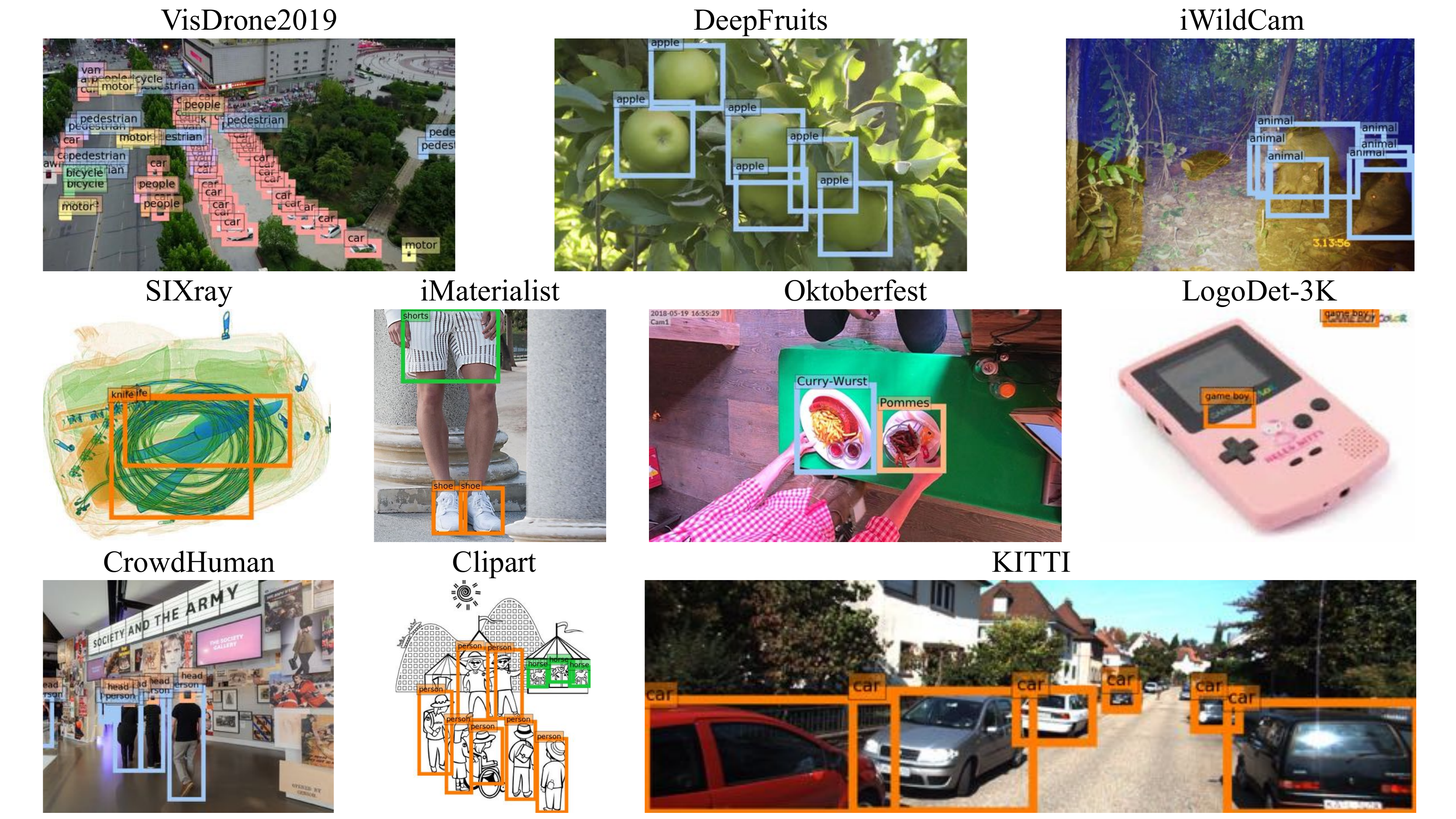}
\caption{Sample images in the proposed FSOD benchmark.}
\label{fig:datasets_0}
\end{figure}

\begin{figure*}[t]
\centering
\begin{subfigure}[t]{0.45\linewidth}
\includegraphics[width=\linewidth]{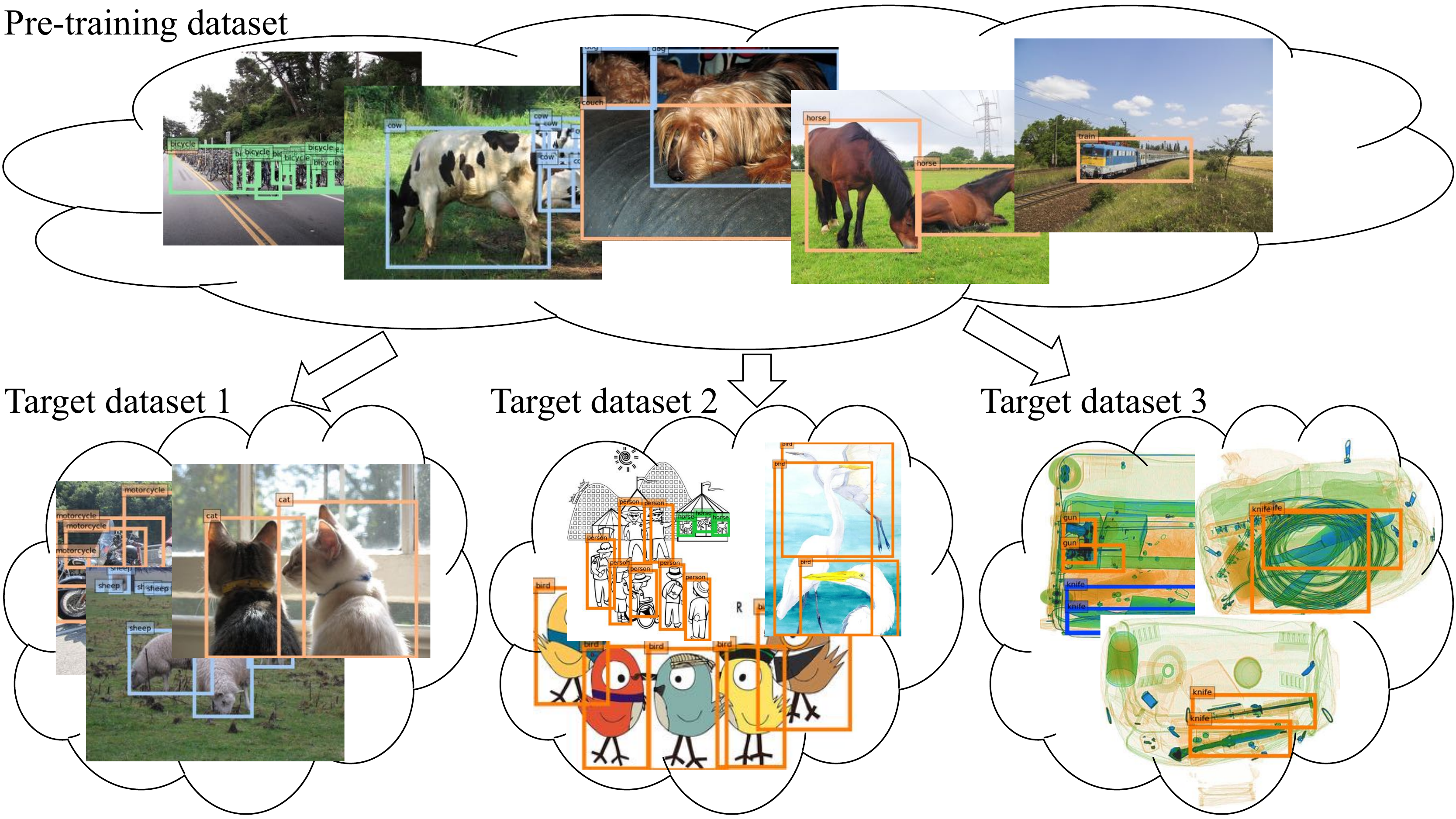}
\caption{Multi-domain datasets.}
\label{fig:cross_domain}
\end{subfigure}
\hspace{\fill}
\begin{subfigure}[t]{0.45\linewidth}
\includegraphics[width=\linewidth]{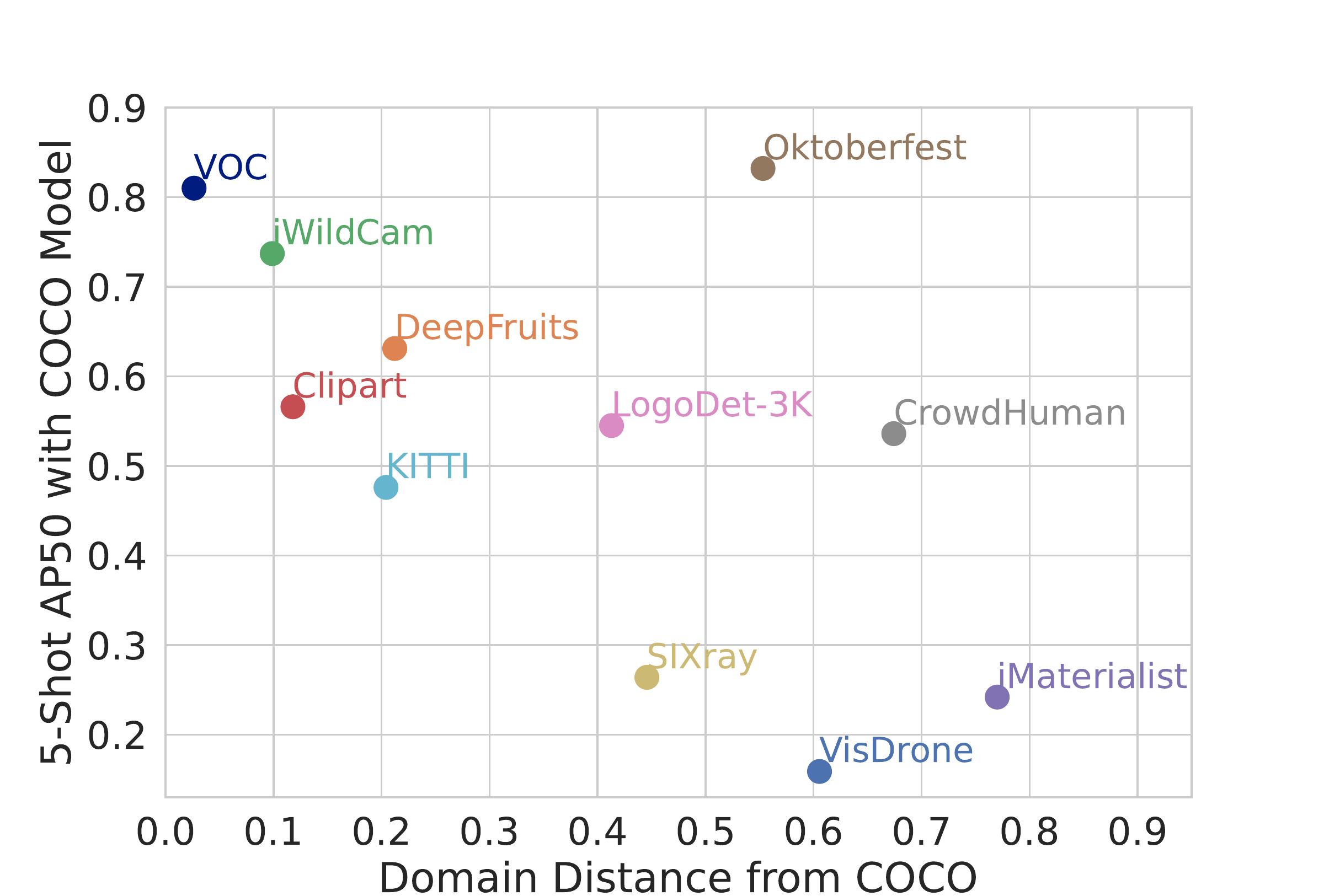}
\caption{Domain distance vs 5-shot performance.}
\label{fig:domain_dis}
\end{subfigure}
\caption{
(a) Real-world applications of FSOD are not limited to the natural image domain; we propose to pre-train models on large-scale natural image datasets and transfer to target domains for few-shot learning. (b) We measure the domain distance between datasets (see \Cref{sec:mofsod} for details) in the benchmark and COCO, and plot against 5-shot AP50 of these datasets fine-tuned from a model pre-trained on COCO. VOC is added for reference. The figure shows the benchmark covers a wide range of domains.
}
\label{fig:motivation}
\end{figure*}

Inspired by Meta-dataset~\cite{triantafillou2020meta}, we propose a Multi-dOmain FSOD (MoFSOD) benchmark consisting of 10 datasets from 10 different domains, as shown in \Cref{fig:datasets_0}. The diversity of MoFSOD datasets can be seen in \Cref{fig:motivation} where the domain distance of each dataset to COCO~\cite{COCO} is depicted. Our benchmark enables us to estimate the performance of FSOD algorithms across domains and settings and helps in better understanding of various factors, such as pre-training, architectures, \etc, that influence the algorithm performance. In addition, we propose a simple natural $K$-shot sampling algorithm that encourages more diversity in class distributions than balanced sampling.

Building on our benchmark, we extensively study the impact of freezing parameters, detection architectures, pre-training datasets, and the effectiveness of several state-of-the-art (SOTA) FSOD algorithms~\cite{wang2020frustratingly,sun2021fsce,qiao2021defrcn}. Our empirical study leads to rethinking conventions in the field and interesting findings that can guide future research on FSOD. 

Conventionally, in FSOD or general few-shot learning, it is believed that freezing parameters to avoid overfitting is helpful or even crucial for good performance~\cite{wang2020frustratingly,sun2021fsce,snell2017proto,wang2019meta}. If we choose to tune more parameters, specific components or designs must be added, such as weight imprinting~\cite{wu2020meta} or decoupled gradient~\cite{qiao2021defrcn}, to prevent overfitting. Our experiments in the MoFSOD show that these design choices might be helpful when pre-training and few-shot learning are in similar domains, as in previous benchmarks. However, if we consider a broader spectrum of domains, unfreezing the whole network results in better overall performance, as the network has more freedom to adapt. We further demonstrate a correlation between the performance gain of tuning more parameters and domain distance (see \Cref{fig:domain_ft}). Overall, \emph{fine-tuning (FT) is a strong baseline} for FSOD on MoFSOD without any bells and whistles.

Using FT as a baseline allows us to explore the impact of different architectures on FSOD tasks. Previous FSOD methods~\cite{wang2020frustratingly,fan2020fsod,sun2021fsce,qiao2021defrcn} need to make architecture-specific design choices; hence focus on a single architecture -- mostly Faster R-CNN~\cite{ren2015faster}, while we conduct extensive study on the impact of different architectures, \eg, recent development of anchor-free~\cite{zhou2021centernet2,zhou2019centernet} and transformer-based architectures~\cite{carion2020end,zhu2020deformable}, on few-shot performance. Surprisingly, we find that even with similar performance on COCO, different architectures have very different downstream few-shot performances. This finding suggests the potential benefits of specifically designed few-shot architectures for improved performance.

Moreover, unlike previous benchmarks, which split the pre-training and few-shot test sets from the same datasets (VOC or COCO), MoFSOD allows us to freely choose different pre-training datasets and explore the potential benefits of large-scale pre-training. To this end, we systematically study the effect of pre-training datasets with ImageNet~\cite{deng2009imagenet}, COCO~\cite{COCO}, LVIS~\cite{lvis}, FSOD-Dataset~\cite{fan2020fsod}, Unified~\cite{zhou2021uni}, and the integration of large-scale language-vision models. Similar to observations in recent works in image classification~\cite{bit} and NLP~\cite{brown2020bert,clip}, we find that large-scale pre-training can play a crucial role for downstream few-shot tasks.

Finally, motivated by the effectiveness of the unfreezing parameters and language-vision pre-training, we propose two extensions: FSCE+ and LVIS+. FSCE+ extends FSCE~\cite{sun2021fsce} to fine-tune more parameters with a simplified fully-connected (FC) detection head. LVIS+ follows the idea of using CLIP embedding of class names as the classifier, but instead of using it in zero-shot/open-vocabulary setting as in~\cite{zhou2021detic,gu2021zero}, we extend it to few-shot fine-tuning. Both methods achieve SOTA results with/without extra pre-training data.

We summarize our contributions as follows:
\begin{itemize}
\item We propose a Multi-dOmain Few-Shot Object Detection (MoFSOD) benchmark to simulate real-world applications in diverse domains.
\item We conduct extensive studies on the effect of architectures, pre-training datasets, and hyperparameters with fine-tuning and several SOTA methods on the proposed benchmark.
We summarize the observations below:
\begin{itemize}
\item \textbf{Unfreezing more layers} do not lead to detrimental overfitting and improve the FSOD performance across different domains.
\item \textbf{Object detection model architectures} have a significant impact on the FSOD performance even when the architectures have a similar performance on the pre-training dataset.
\item \textbf{Pre-training datasets} play an important role in the downstream FSOD performance. Effective utilization of the pre-training dataset can significantly boost performance.
\end{itemize}
\item Based on these findings, we propose two extensions that outperform SOTA methods by a significant margin on our benchmark.
\end{itemize}

%% file: 2_related.tex
\section{Related Work} \label{sec:related}
\textbf{Meta-learning-based methods} for FSOD are inspired by few-shot classification.
Kang et al.~\cite{kang2019few} proposed a meta feature extractor with feature reweight module, which maps support images to mean features and reweight query features with the mean features, inspired by protoypical networks~\cite{snell2017proto}. Meta R-CNN~\cite{yan2019meta} extended the idea with an extra predictor-head remodeling network to extract class-attentive vectors. MetaDet~\cite{wang2019meta} proposed meta-knowledge transfer with weight prediction module.

\textbf{Two-stream methods} take one query image and support images as inputs, and use the correlations between query and support features as the final features to the detection head and the Region Proposal Network (RPN). Several works in this direction~\cite{fan2020fsod,zhang2021,han2021iccv} have shown competitive results. These methods require all classes to have at least one support image to be fed to the model, which makes the overall process slow.

\textbf{Fine-tuning-based methods} update only the linear classification and regression layers~\cite{wang2020frustratingly}, the whole detection head and RPN with an additional contrastive loss~\cite{sun2021fsce}, or decoupling the gradient of RPN and the detection head while updating the whole network~\cite{qiao2021defrcn}. These methods are simple yet have shown competitive results. We focus on benchmarking them due to their simplicity, efficiency, and higher performance than other types.

\textbf{Multi-domain few-shot classification benchmarks.} In few-shot classification, miniImageNet~\cite{vinyals16mini} and tieredImageNet~\cite{ren2018tiered} have been used as standard benchmarks. Similar to benchmarks in FSOD, they are divided into two splits and used for pre-training and few-shot learning, respectively, such that they are in the same natural image domain. Recent works have proposed new benchmarks to address this issue: Tseng~\etal~\cite{tseng2020cross} proposed a cross-domain few-shot classification benchmark with five datasets from different domains. Triantafillou~\etal~\cite{triantafillou2020meta} proposed Meta-Dataset, which is a large-scale few-shot classification benchmark with ten datasets and a sophisticatedly designed sampling algorithm to sample realistically imbalanced few-shot training datasets. Although not specifically catering to few-shot applications, Wang~\etal~\cite{wang2019towards} proposed universal object detection, which aims to cover multi-domain datasets with a single model for high-shot object detection.

%% file: 3_fsod.tex
\section{MoFSOD: A Multi-Domain FSOD Benchmark}
In this section, we first describe existing FSOD benchmarks and their limitations. Then, we propose a Multi-dOmain FSOD (MoFSOD) benchmark.

\subsection{Existing Benchmarks and Limitations}

Recent FSOD works have evaluated their methods in PASCAL VOC $15+5$ and MS COCO $60+20$ benchmarks proposed by Kang~\etal~\cite{kang2019few}. From the original VOC~\cite{VOC} with 20 classes and COCO~\cite{COCO} with 80 classes,  they took 25\% of classes as novel classes for few-shot learning, and the rest of them as base classes for pre-training. For VOC $15+5$, three splits were made, where each of them consists of $15$ base classes for pre-training, and the other 5 novel classes for few-shot learning. For each novel class, $K = \{1, 3, 5, 10\}$ object instances are sampled, which are referred to as shot numbers. For COCO, 20 classes overlapped with VOC are considered novel classes, and $K=\{10, 30\}$-shot settings are used. Different from classification, as an image usually contains multiple annotations in object detection, sampling exactly $K$ annotations per class is difficult. Kang~\etal~\cite{kang2019few} proposed pre-defined support sets for few-shot training, which would cause overfitting~\cite{huang2021survey}. Wang~\etal~\cite{wang2020frustratingly} proposed to sample few-shot training datasets with different random seeds to mitigate this issue, but the resulting sampled datasets often contain more than $K$ instances. While these benchmarks contributed to the research progress in FSOD, they have several limitations.

First, these benchmarks do not capture the breadth of few-shot tasks and domains as they sample few-shot task instances from a single dataset, as we discussed in \Cref{sec:intro}. Second, these benchmarks contain only a fixed number of classes, 5 or 20.
However, real-world applications might have a varying number of classes, ranging from one class, \eg, face/pedestrian detection~\cite{widerface,widerperson}, to thousands of classes, \eg, logo detection~\cite{logodet3k}. Last but not least, these benchmarks are constructed with the balanced $K$-shot sampling. For example, in the 5-shot setting, a set of images containing exactly 5 objects~\cite{kang2019few} is pre-defined. Such a setting is unlikely in real-world few-shot tasks.
We also demonstrate that such a sampling strategy can lead to high variances in the performance of multiple episodes (see \Cref{tab:sota_balanced}). Moreover, different from classification, object detection datasets tend to be imbalanced due to the multi-label nature of the datasets. For example, COCO~\cite{COCO} and OpenImages~\cite{OpenImages} have a more dominant number of person instances than any other objects. The benchmark datasets should also explore these imbalanced scenarios.

\subsection{Multi-Domain Benchmark Datasets}
\label{sec:mofsod}

While FSOD applications span a wide range of domains, gathering enough pre-training data from these domains might be difficult. Hence, it becomes important to test the few-shot algorithm performance in settings where the pre-training and few-shot domains are different. Similar to Meta-dataset~\cite{triantafillou2020meta} in few-shot classification, we propose to extend the benchmark with datasets from a wide range of domains rather than a subset of natural image datasets. Our proposed benchmark consists of 10 datasets from 10 domains:
VisDrone~\cite{visdrone} in aerial images,
DeepFruits~\cite{deepfruits} in agriculture,
iWildCam~\cite{iwildcam} in animals in the wild,
Clipart in cartoon,
iMaterialist~\cite{imaterialist} in fashion,
Oktoberfest~\cite{oktoberfest} in food,
LogoDet-3K~\cite{logodet3k} in logo,
CrowdHuman~\cite{crowdhuman} in person,
SIXray~\cite{sixray} in security, and
KITTI~\cite{kitti} in traffic/autonomous driving.
We provide statistics of these datasets in \Cref{tab:stats}. The number of classes varies from 1 to 352, and that of boxes per image varies from 1.2 to 54.4, covering a wide range of scenarios.

\input{table_pretraining}

In \Cref{fig:domain_dis}, we illustrate the diversity of domains in our benchmark by computing the domain distances between these datasets and COCO~\cite{COCO} and plotting against the 5-shot performance of fine-tuning (FT) on each dataset from a model pre-trained on COCO. Specifically, we measure the domain similarity by calculating the recall of a pre-trained COCO model on each dataset in a class-agnostic fashion, similar to the measurement of unsupervised object proposals~\cite{hosang2014proposal}.
Intuitively, if a dataset is in a domain similar to COCO, then objects in the dataset are likely to be localized well by the model pre-trained on COCO. As a reference, VOC has a recall of 97\%. For presentation purpose, we define $(1 - \text{recall})$ as the domain distance. We can see diverse domain distances in the benchmark, ranging from 0.1 to 0.8. 
Interestingly, the domain distance also correlates with the FSOD performance. Although this is not the only deciding factor, as the intrinsic properties (such as the similarity between training and test datasets) of a dataset also play an important role, we can still see the linear correlation between the domain distance and 5-shot performance with the Pearson correlation coefficient -0.43. Oktoberfest and CrowdHuman are outliers in our analysis possibly as they are relatively easy.

\textbf{Natural $K$-Shot Sampling.}
We use a natural $K$-shot sampling algorithm to maintain the original class distribution for this benchmark. Specifically,
we sample $C \times K$ images from the original dataset without worrying about class labels, where $C$ is the number of classes of the original dataset. Then, we check missing images to ensure we have at least one image for each class of all classes. We provide the details in \appendixa. The comparison between the balanced $K$-shot and natural $K$-shot sampling shows that our conclusions do not change based on the sampling algorithm, but the performance of the natural $K$-shot sampling is more consistent on different episodes (see \Cref{tab:sota_balanced}) and covers imbalanced class distributions existing in the real-world applications.

\textbf{Evaluation Protocol.}
To evaluate the scalability of methods, we experiment with four different average shot numbers, $K = \{1, 3, 5, 10\}$. We first sample a few-shot training dataset from the original training dataset with the natural $K$-shot sampling algorithm for each episode. Then, we initialize the object detection model with pre-trained model parameters and train the model with FSOD methods. For evaluation, we randomly sample 1k images from the original test set if the test set is larger than 1k. We repeat this episode 10 times with different random seeds for all multi-domain datasets and report the average of the mean and standard deviation of the performance.

\textbf{Metrics.}
As evaluation metrics, we use AP50 and the average rank among compared methods~\cite{triantafillou2020meta}. AP50 stands for the average precision of predictions where the IoU threshold is 0.5, and the rank is an integer ranging from 1 to the number of compared methods, where the method with the highest AP50 gets rank 1. We first take the best AP50 among different hyperparameters at the end of training for each episode, compute the mean and standard deviation of AP50 and the rank over 10 episodes, and then average them over different datasets and/or shots, depending on the experiments.

%% file: table_pretraining.tex
\begin{table}[t]
\caption{Statistics of pre-training and MoFSOD datasets (\Cref{tab:stats}) and performance of different architectures on the pre-training datasets (\Cref{tab:pre-perf}).}
\label{tab:table1}
\begin{subtable}[t]{0.54\textwidth}
\centering
\captionsetup{font=scriptsize}
\resizebox{\linewidth}{!}{
\begin{tabular}[t]{ccccc}
\toprule
Domain & Dataset & \# classes & \# training images & \cr
\cmidrule(rl){1-1}\cmidrule(rl){2-2}\cmidrule(rl){3-3}\cmidrule(rl){4-4}
& COCO & 80 & 117k \cr
\rowcolor{Gray} \cellcolor{White} & FSODD & 800 & 56k \cr
& LVIS &  1203 & 100k \cr
\rowcolor{Gray} \cellcolor{White} \multirowcell{-4}{Natural\\Image} & Unified & 723  & 2M \cr
\bottomrule \toprule
\rowcolor{White} Domain & Dataset & \# classes & \# bboxes per image \cr
\cmidrule(rl){1-1}\cmidrule(rl){2-2}\cmidrule(rl){3-3}\cmidrule(rl){4-4}
\rowcolor{White} Aerial & VisDrone & 10 & 54.4 \cr
\rowcolor{Gray} Agriculture & DeepFruits & 7 & 5.6 \cr
\rowcolor{White} Animal & iWildCam & 1 & 1.5  \cr
\rowcolor{Gray} Cartoon & Clipart & 20 & 3.3  \cr
\rowcolor{White} Fashion & iMaterialist & 46  & 7.3 \cr
\rowcolor{Gray} Food & Oktoberfest & 15 & 2.4  \cr
\rowcolor{White} Logo & LogoDet-3K & 352 & 1.2  \cr
\rowcolor{Gray} Person & CrowdHuman & 2 & 47.1  \cr
\rowcolor{White} Security & SIXray & 5 & 2.1  \cr
\rowcolor{Gray} Traffic & KITTI & 4 & 7.0  \cr
\bottomrule
\end{tabular}
}
\caption{{\bf Top:} statistics of pre-training datasets.\\\hspace*{13pt}{\bf Bottom:} statistics of MoFSOD datasets.}
\label{tab:stats}
\end{subtable}
\hspace{\fill}
\begin{subtable}[t]{0.44\textwidth}
\centering
\captionsetup{font=scriptsize}
\begin{tabular}[t]{ccc}
\toprule
Dataset & Architecture & AP \cr
\cmidrule(rl){1-1}\cmidrule(rl){2-2}\cmidrule(rl){3-3}
 & Faster R-CNN & $42.7$ \cr
\rowcolor{Gray} \cellcolor{White} & Cascade R-CNN & $45.1$ \cr
 & CenterNet2 & $45.3$ \cr
\rowcolor{Gray} \cellcolor{White} & RetinaNet & $39.3$ \cr
 & Deformable DETR & $46.3$ \cr
\rowcolor{Gray} \cellcolor{White} \multirow{-6}{*}{COCO~} & Cascade R-CNN-P67~ & $45.9$ \cr
\midrule
& Faster R-CNN & $24.2$ \cr
\rowcolor{Gray} \cellcolor{White} & CenterNet2 & $28.3$ \cr
\multirow{-3}{*}{LVIS~} & Cascade R-CNN-P67~ & $26.2$ \cr
\bottomrule
\end{tabular}
\caption{\scriptsize{Performance of benchmark architectures pre-trained on COCO and LVIS.
FSODD and Unified do not have a pre-defined validation/test set, so we do not measure their pre-training performances.}}
\label{tab:pre-perf}
\end{subtable}
%
%
\end{table}

%% file: 4_exp.tex
\section{Experiments}
\label{sec:exp}
In this section, we conduct extensive experiments on MoFSOD and discuss the results.
For better presentations, we highlight compared methods and architectures in \textit{italics} and pre-training/few-shot datasets in \textbf{bold}.
\subsection{Experimental Setup}

\textbf{Model architecture.}
We conduct experiments on six different architectures. For simplicity, we use ResNet-50~\cite{he2016deep} as the backbone of all architectures. We also employ deformable convolution v2~\cite{zhu19deconvv2} in the last three stages of the backbone. Specifically, we benchmark two-stage object detection architectures: 1) \textit{Faster R-CNN}~\cite{ren2015faster} 2) \textit{Cascade R-CNN}~\cite{cai2018cascade}, and the newly proposed 3) \textit{CenterNet2}~\cite{zhou2021centernet2},
and one-stage architectures: 4) \textit{RetinaNet}~\cite{lin2017retina}, as well as transformer-based 5) \textit{Deformable-DETR}~\cite{zhu2020deformable}. Note that all architectures utilize Feature Pyramid Networks (FPN)~\cite{lin2017feature} or similar multi-scale feature aggregation techniques. In addition, we also experiment the combination of the FPN-P67 design from \textit{RetinaNet} and \textit{Cascade R-CNN}, dubbed 6) \textit{Cascade R-CNN-P67}~\cite{zhou2021uni}. We conduct our architecture analysis pre-trained on \textbf{COCO}~\cite{COCO} and \textbf{LVIS}~\cite{lvis}. \Cref{tab:pre-perf} summarizes the architectures and their pre-training performance.

\textbf{Freezing parameters.}
Based on the design of detectors, we can think of three different levels of fine-tuning the network:
1)~only the last classification and regression fully-connected (FC) layer~\cite{wang2020frustratingly},
2)~the detection head consisting of several FC and/or convolutional layers~\cite{fan2020fsod}, and
3)~the whole network, \ie, standard fine-tuning.\footnote{%
When training an object detection model, the batch normalization layers~\cite{ioffe2015batch} and the first two macro blocks of the backbone (\texttt{stem} and \texttt{res2}) are usually frozen, even for large-scale datasets. We follow this convention in our paper.
}
We study the effects of these three ways of tuning on different domains in MoFSOD with \textit{Faster R-CNN}~\cite{ren2015faster}.

\textbf{Pre-training datasets.}
To explore the effect of pre-training dataset, we conduct experiments on five pre-training datasets: \textbf{ImageNet}\footnote{%
This is ImageNet-1K for classification, which is commonly used for pre-training standard object detection methods, \ie, we omit pre-training on an object detection task.%
}~\cite{deng2009imagenet}, \textbf{COCO}~\cite{COCO}, \textbf{FSODD}\footnote{%
The name of the dataset is also FSOD, so we introduce an additional D to distinguish the dataset from the task.%
}~\cite{fan2020fsod}, \textbf{LVIS}~\cite{lvis}, and \textbf{Unified}, which is a union of OpenImages v5~\cite{OpenImages}, Object365 v1~\cite{O365}, Mapillary~\cite{mapillary}, and \textbf{COCO}, combined as in~\cite{zhou2021uni}. To reduce the combinations of different architectures and pre-training datasets, we conduct most of the studies on the best performing architecture \textit{Cascade R-CNN-P67}. Also, inspired by \cite{zhou2021detic,gu2021zero}, we experiment with the effect of CLIP~\cite{clip} embeddings to initialize the final classification layer of detector when pre-training on \textbf{LVIS} dubbed \textbf{LVIS+}. In addition, \textbf{LVIS++} uses the backbone pre-trained on the ImageNet-21K classification task instead of ImageNet-1K before pre-training on \textbf{LVIS}. All experiments are done with Detectron2~\cite{wu2019detectron2}.

\textbf{Hyperparameters.}
For pre-training, we mostly follow standard hyperparameters of the corresponding method, with the addition of deformable convolution v2~\cite{zhu19deconvv2}.
On \textbf{COCO} and \textbf{LVIS},
for \textit{Faster R-CNN}, \textit{Cascade R-CNN}, and \textit{Cascade R-CNN-P67}, we use the 3$\times$ scheduler with 270k iterations, the batch size of 16, the SGD optimizer with initial learning rate of 0.02 decaying by the factor of 0.1 at 210k and 250k.
For \textit{RetinaNet}, the initial learning rate is 0.01~\cite{lin2017retina}.
For \textit{CenterNet2}, following~\cite{zhou2021centernet2}, we use the \textit{CenterNet}~\cite{zhou2019centernet} as the first stage and the Cascade R-CNN head as the second stage, where the other hyperparameters are the same as above.
For \textit{Deformable-DETR}~\cite{zhu2020deformable}, we follow the two-stage training of 50 epochs, the AdamW optimizer, and the initial learning rate of 0.0002 decaying by the factor of 0.1 at 40 epochs. On \textbf{FSODD}, we train for 60 epochs with the learning rate of 0.02 decaying by the factor of 0.1 at 40 and 54 epochs and the batch size of 32. On \textbf{Unified}, following~\cite{zhou2021uni}, the label space of four datasets are unified, the dataset-aware sampling and equalization loss~\cite{tan2020eql} are applied to handle long-tailed distributions, and training is done for 600k iterations with the learning rate of 0.02 decaying by the factor of 0.1 at 400k and 540k iterations, and the batch size of 32.

For few-shot training, we train models for 2k iterations with the batch size of 4 on a single V100 GPU.\footnote{The batch size could be less than 4 if the sampled dataset size is less than 4, \eg, when the number of classes and shot number $K$ is 1, and the batch size has to be 1.}
For \textit{Faster R-CNN}, \textit{Cascade R-CNN}, \textit{Cascade R-CNN-P67}, and \textit{CenterNet2}, we train models with the SGD optimizer and different initial learning rates in \{0.0025, 0.005, 0.01\} and choose the best, where the learning rate is decayed by the factor of 0.1 after 80\% of training.
For \textit{RetinaNet}, we halve the learning rates to \{0.001, 0.0025, 0.005\}, as we often observe training diverges with the learning rate of 0.01. For \textit{Deformable-DETR} and \textit{CenterNet2 with CLIP}, we use the AdamW optimizer and initial learning rates in \{0.0001, 0.0002, 0.0004\}.\footnote{%
The learning rates are chosen from our initial experiments on three datasets. Note that training \textit{CenterNet2} with AdamW results in worse performance than SGD.%
}

\textbf{Compared methods.}
\textit{TFA~\cite{wang2020frustratingly}} or \textit{Two-stage Fine-tuning Approach} has shown to be a simple yet effective method for FSOD. 
TFA fine-tunes the box regressor and classifier on the few-shot dataset while freezing the other parameters of the model.
For this method, we use the FC head, such that TFA is essentially the same as tuning the final FC layer.\footnote{%
Replacing the FC head with the cosine-similarity results in a similar performance.}

\textit{FSCE~\cite{sun2021fsce}} or \textit{Few-Shot object detection via Contrastive proposals Encoding} improves TFA by 1)~additionally unfreezing detection head in the setting of TFA, 2)~doubling the number of proposals kept after NMS and halving the number of sampled proposals in the RoI head, and 3)~optimizing the contrastive proposal encoding loss. While the original work did not apply the contrastive loss for extremely few-shot settings (less than 3), we explicitly compare two versions in all shots: without (\textit{FSCE-base}, the same as tuning the detection head) and with the contrastive loss (\textit{FSCE-con}).

\textit{DeFRCN~\cite{qiao2021defrcn}} or \textit{Decoupled Faster R-CNN} can be distinguished with other methods by 1)~freezing only the R-CNN head, 2)~decoupling gradients to suppress gradients from RPN while scaling those from the R-CNN head, and 3)~calibrating the classification score from an offline prototypical calibration block (PCB), which is a CNN-based prototype classifier pre-trained on ImageNet~\cite{deng2009imagenet}. We note that PCB does not re-scale input images in their original implementation, unlike the object detector, so we manually scaled images to avoid GPU memory overflow if they are too large.

\textit{FT} or \textit{Fine-tuning}
does not freeze model parameters as done for other methods. Though it is undervalued in prior works, we found that this simple baseline outperforms state-of-the-art methods in our proposed benchmark. All experiments are done with this method unless otherwise specified.
\input{table_overall}

\subsection{Experimental Results and Discussions}
\label{exp:analysis}

\textbf{Effect of tuning more parameters.}
We first analyze the effect of tuning more or fewer parameters on MoFSOD. In \Cref{tab:sota_natural} and \Cref{tab:tune}, we examine three methods freezing different number of parameters when fine-tuning: \textit{TFA} as tuning the last FC layers only, \textit{FSCE-base} as tuning the detection head, and  \textit{FT} as tuning the whole network. We observe that freezing fewer parameters improves the average performance: tuning the whole network (\textit{FT}) shows better performance than others, while tuning the last FC layers only (\textit{TFA}) shows lower performance than others. Also, the performance gap becomes larger as the size of few-shot training datasets increases. For example, \textit{FT} outperforms \textit{FSCE-base} and \textit{TFA} by 1.0\% and 8.1\% in 1-shot, and 2.4\%, 17.4\% in 10-shot, respectively.
This contrasts with the conventional belief that freezing most of the parameters generally improves the performance of few-shot learning, as it prevents overfitting~\cite{snell2017proto,finn2017maml,wang2020frustratingly,sun2021fsce}. However, this is not necessarily true for FSOD. For example, in the standard two-stage object detector training, RPN is class-agnostic, such that its initialization for training downstream few-shot tasks can be the one pre-trained on large-scale datasets, preserving the pre-trained knowledge on objectness. Also, the detection head utilizes thousands of examples even in few-shot scenarios, because RPN could generate 1--2k proposals per image. Hence, the risk of overfitting is relatively low.

\begin{figure*}[t]
\centering
\begin{subfigure}[t]{0.47\linewidth}
\includegraphics[width=\linewidth]{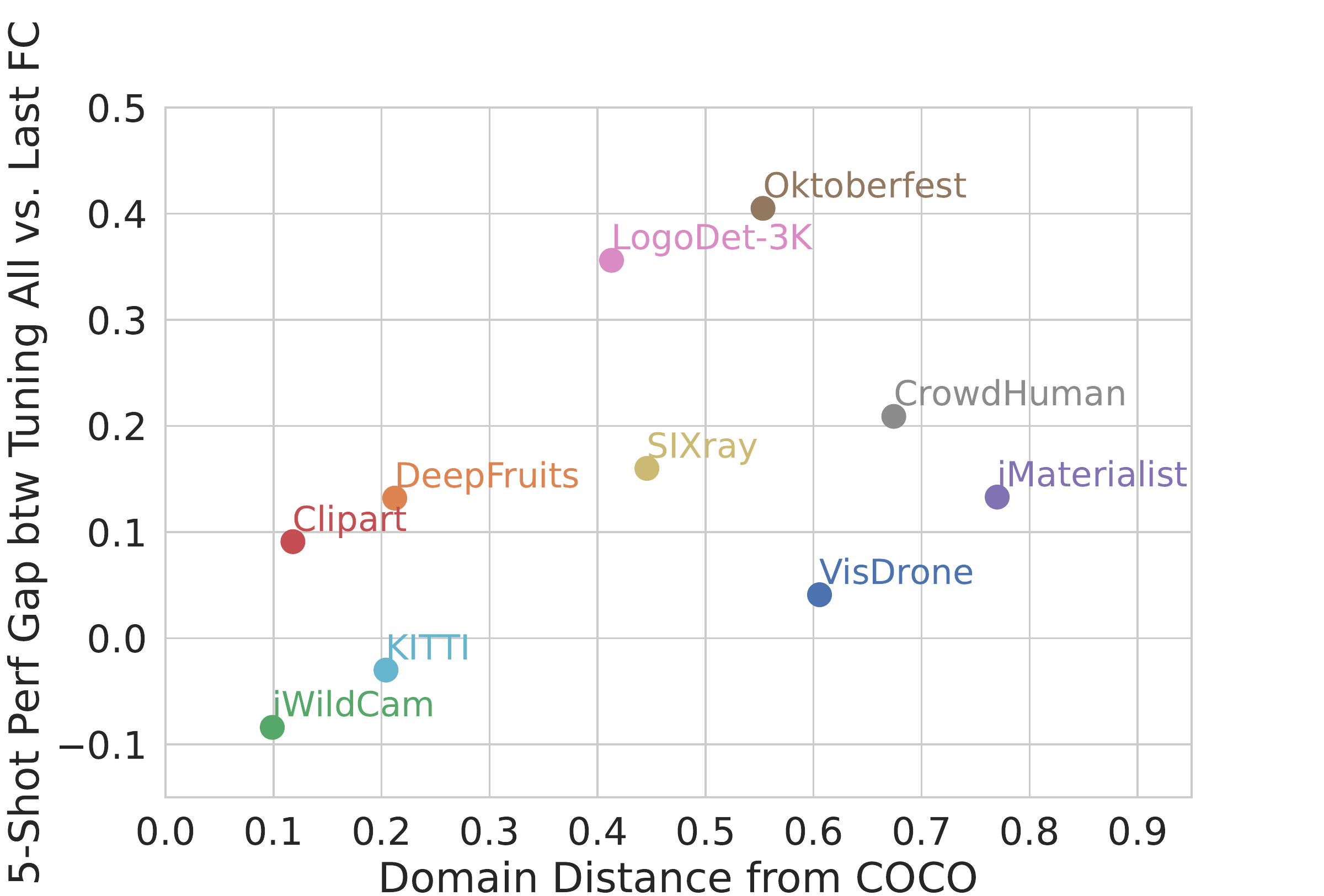}
\caption{Performance gain by tuning \textit{the whole network} \textit{FT}) rather than \textit{the last FC layer only} (\textit{TFA}) vs. domain distance.}
\label{fig:domain_ft}
\end{subfigure}
\hspace{\fill}
\begin{subfigure}[t]{0.45\linewidth}
\includegraphics[width=\linewidth]{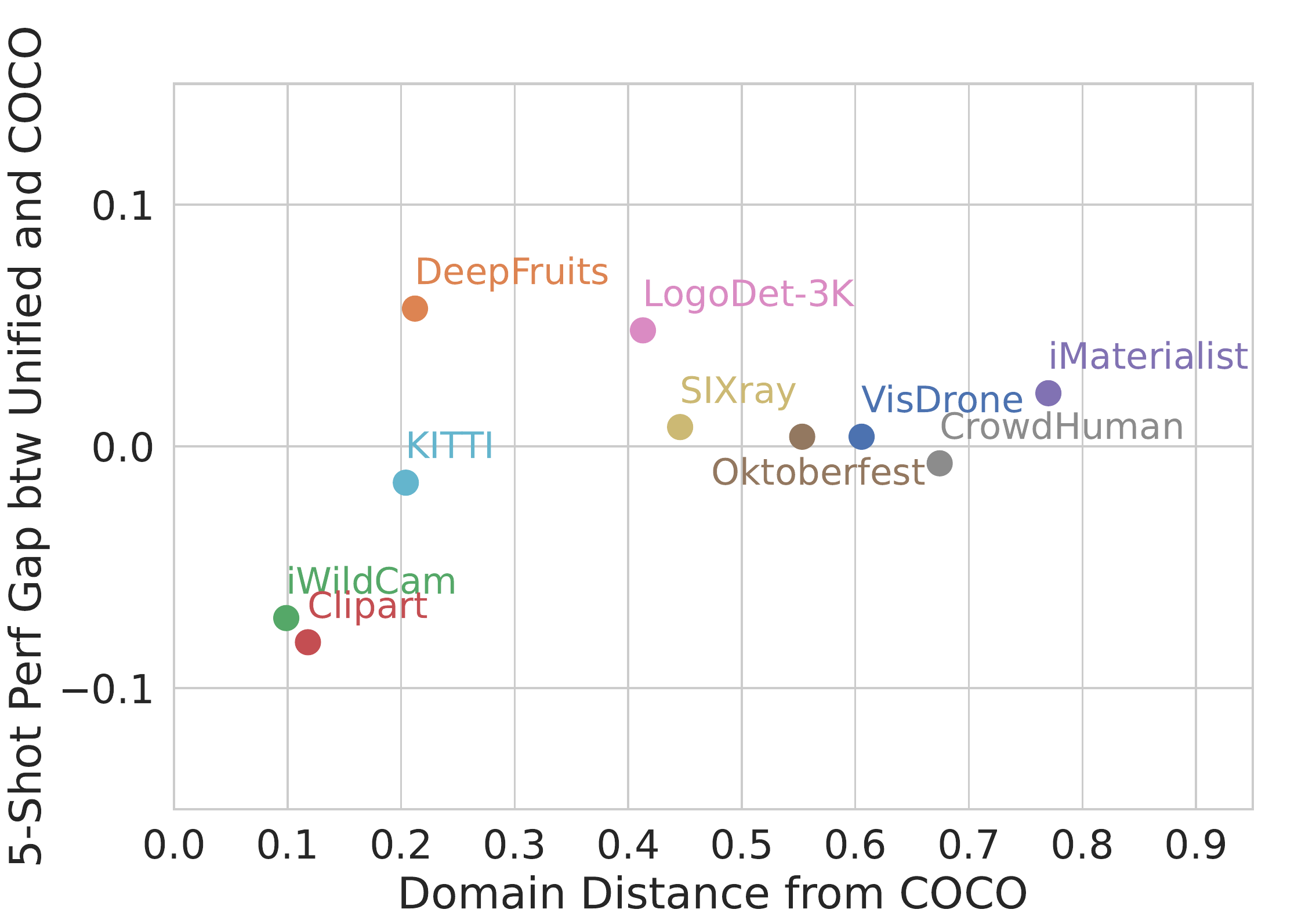}
\caption{Performance gap between \textbf{Unified} and \textbf{COCO} vs. domain distance.}
\label{fig:domain_pre}
\end{subfigure}
\caption{We demonstrate the correlation between tuning more parameters and domain distance in \Cref{fig:domain_ft} and the correlation between pre-training datasets and domain distance in \Cref{fig:domain_pre}.}
\label{fig:domain_dist}
\end{figure*}

However, fine-tuning more parameters does not always improve performance. \Cref{fig:domain_ft} illustrates the performance gain by tuning more parameters with respect to the domain distance. There is a linear correlation, \ie, the performance gain by fine-tuning more parameters increases when the domain distance increases. This implies that fine-tuning fewer parameters to preserve the pre-trained knowledge is better when the the few-shot dataset is close to the pre-training dataset. Hence, for datasets close to \textbf{COCO}, such as \textbf{KITTI} and \textbf{iWildCam}, \textit{tuning Last FC layers} (\textit{TFA}) is the best performing method. From these observations, an interesting research direction might be exploring a sophisticated tuning of layers based on the few-shot problem definition and the domain gap between pre-training and few-shot tasks. 

One way to design such sophisticated tuning is to develop a better measure for domain distance. In fact, the proposed measure with class-agnostic object recall has limitations. If we decouple the object detection task into localization (background vs. foreground) and classification (among foreground classes), then the proposed domain distance is biased towards measuring localization gaps. Therefore, it ignores the potential classification gaps that would also require tuning more layers. For example, although we can get a good coverage with the proposals of \textbf{COCO} pre-training for \textbf{Clipart} (classic domain adaption dataset) and \textbf{DeepFruits} (infrared images), resulting in relatively small domain distances, there exist significant gaps of feature discrimination for fine-grained classification. In this case, we need to tune more parameters for better performance.

\input{table_arch_5shot}

\textbf{Effect of model architectures.}
A benefit of using \textit{FT} as the baseline is that we can systematically study the effects of model architectures without the constraints of specifically designed components. Many different architectures have been proposed to solve object detection problems; each has its own merits and drawbacks. One-stage methods, such as YOLO~\cite{redmon2017yolo9000} and RetinaNet~\cite{lin2017retina}, are known for their fast inference speed. However, different from two-stage methods, they do not have the benefit of inheriting the pre-trained class-agnostic RPN. Specifically, the classification layers for discriminating background/foreground and foreground classes need to be reinitialized, as they are often tied together in one-stage methods. We validate this hypothesis by comparing with two-stage methods in \Cref{tab:arch_overall}. Compared to \textit{Faster R-CNN}, \textit{RetinaNet} has 4--6\% low performance on MoFSOD. Per-dataset performance in \Cref{tab:arch} shows that \textit{RetinaNet} is worse in all cases.

The same principle of preserving as much information as possible from pre-training also applies for two-stage methods,
\ie, reducing the number of randomly reinitialized parameters is better.
Specifically, we look into the performance of \textit{Cascade R-CNN} vs \textit{Faster R-CNN}. For \textit{Cascade R-CNN}, we need to reinitialize and learn three FC layers as there are three stages in the cascade detection head, while we only need to reinitialize the last FC layer for \textit{Faster R-CNN}. However, \textit{Cascade R-CNN} is known to have better performance,
as demonstrated in \Cref{tab:pre-perf}. In FSOD, these two factors cancelled out, such that their performance is on a par with each other as shown in \Cref{tab:arch_overall}.

Based on these insights, we extend \textit{Cascade R-CNN} by applying the FPN-P67 architecture~\cite{lin2017retina}, similar to \cite{zhou2021uni,zhou2021centernet2}. Specifically, assuming ResNet-like architecture~\cite{he2016deep}, we use the last three stages of the backbone, namely $\left[\texttt{res3}, \texttt{res4}, \texttt{res5}\right]$, instead of the last four in standard FPN. Then, we add P6 and P7 of the FPN features from P5 with two different FC layers, such that RPN takes in features from $\left[\texttt{P3}, \texttt{P4}, \texttt{P5}, \texttt{P6}, \texttt{P7}\right]$ to improve the class-agnostic coverage, which can be inherited for down-stream tasks. While the detection head still uses $\left[\texttt{P3}, \texttt{P4}, \texttt{P5}\right]$ only. As shown in \Cref{tab:arch_overall}, the resulting architecture, \textit{Cascade R-CNN-P67} improves \textit{Cascade R-CNN} by 3--4\% on the downstream few-shot tasks.

Moreover, recent works proposed new directions of improvement, such as utilizing point-based predictions~\cite{zhou2019centernet,zhou2021centernet2} or transformer-based set predictions~\cite{carion2020end,zhu2020deformable}. These methods are unknown quantities in FSOD, as no previous FSOD work has studied them. In our experiments, while \textit{CenterNet2}~\cite{zhou2021centernet2} outperforms \textit{Faster R-CNN} on \textbf{COCO} by 2.6\% as shown in \Cref{tab:pre-perf}, its FSOD performance on MoFSOD is lower, \eg, 2.4\% in 1-shot. In the case of \textit{Deformable-DETR}~\cite{zhu2020deformable}, it outperforms \textit{Faster R-CNN} in both pre-training and few-shot learning, by 3.6\% on \textbf{COCO} and 2.1\% on MoFSOD in 10-shot. These results show that the upstream performance might not necessarily translate to the downstream FSOD performance. We note that we could not observe a significant correlation between the performance gap of different architectures and domain distances.

\textbf{Effect of pre-training datasets.}
MoFSOD consists of datasets from a wide range of domains, allowing us to freely explore different pre-training datasets while ensuring domain shifts between pre-training and few-shot learning. We examine the impact of the pre-training datasets with the best performing \textit{Cascade R-CNN-P67} architecture.

In \Cref{tab:pre_training_overall}, we first observe that pre-training on \textbf{ImageNet} for classification results in low performance, as it does not provide a good initialization for downstream FSOD, especially for RPN. On the other hand, compared to \textbf{COCO}, \textbf{FSODD}, \textbf{LVIS}, and \textbf{Unified} have more classes and/or more annotations, while they have a fewer, similar, and more number of images, respectively. Pre-training on these larger object detection datasets does not improve the FSOD performance significantly, as shown in \Cref{tab:pre_training_overall}. For example, pre-training on \textbf{Unified} improves the performance over \textbf{COCO} when the domain distance from \textbf{COCO} is large, such as \textbf{Deepfruits} and \textbf{LogoDet-3K} as shown in \Cref{fig:domain_pre}.
However, pre-training on \textbf{Unified} results in lower performance for few-shot datasets close to \textbf{COCO}, such that the overall performance is similar. We hypothesize that this could be due to the non-optimal pre-training of \textbf{LVIS} and \textbf{Unified}, as these two datasets are highly imbalanced and difficult to train. It could also be the case that even \textbf{LVIS} and \textbf{Unified} do not have better coverage for these datasets.

On the other end of the spectrum, we can combine the idea of preserving knowledge and large-scale pre-training by utilizing a large-scale language-vision model. Following~\cite{gu2021zero,zhou2021detic}, we use CLIP to extract text features from each class name and build a classifier initialized with the text features.
In this way, we initialize the classifier with the CLIP text embeddings for downstream few-shot tasks, such that it has strong built-in knowledge of text-image alignment, better than random initialization.
As demonstrated in \Cref{tab:pre_training_overall}, For \textbf{LVIS+}, we can see this improves performance significantly by 7.5\% for \textit{CenterNet2}, and 3.3\% for \textit{Cascade R-CNN-P67} in 5-shot. \textbf{LVIS++} pre-trains the backbone on ImageNet-21K instead of ImageNet-1K (before pre-training on \textbf{LVIS}), and it further improves over \textbf{LVIS+} by 0.8\% in 5-shot. 
However, the benefit of CLIP initialization is valid only when class names are matched with texts presented in CLIP; an exceptional case is \textbf{Oktoberfest}, which has German class names, such that \textbf{LVIS+} does not help.

\textbf{Comparison with SOTA methods.}
\Cref{tab:sota_natural} and \Cref{tab:sota_balanced} compare SOTA methods and our proposed methods.
For balanced $K$-shot sampling, we follow Wang~\etal~\cite{wang2020frustratingly} to sample $K$ instances for each class whenever possible greedily. Here \textit{FT} and \textit{FSCE+} employ a similar backbone/architecture and pre-trained data as all SOTA methods for a fair comparison. We confirm that \textit{FT} is indeed a strong baseline, such that it performs better than other SOTA methods in both natural $K$-shot and balanced $K$-shot settings.
In addition to \textit{FT}, based on the insights above, we propose several extensions: 
1)~\textit{FSCE+} is an extension of \textit{FSCE} by tuning the whole network parameters, similar to \textit{FT}. We keep the contrastive proposal encoding loss, but we simplify the classification head from the cosine similarity head to the FC head. We can see the improvement by 2--3\% compared to \textit{FSCE} for both natural $K$-shot and balanced $K$-shot scenarios and performs slightly better than \textit{FT}.
2)~\textit{FT+} replaces \textit{Faster R-CNN} with \textit{Cascade R-CNN-P67}, and it improves over \textit{FT} by 3--4\% without sacrificing inference speed or memory consumption much.
3)~\textit{FT++} replaces \textit{Faster R-CNN} with \textit{CenterNet2} and uses \textbf{LVIS++} for initialization, and it further improves the performance by around 3\% in 3-, 5-, and 10-shot.
We also observe that while the overall trend of performance is similar for both natural and balanced $K$-shot sampling, the standard deviation of the natural $K$-shot performance is less than that of the balanced $K$-shot.

%% file: table_overall.tex
\begin{table*}[t]
\def\smidrule{\cmidrule(rl){1-1}\cmidrule(rl){2-2}\cmidrule(rl){3-3}\cmidrule(rl){4-4}\cmidrule(rl){5-5}\cmidrule(rl){6-6}}
\centering
\caption{The $K$-shot performance on MoFSOD with $K = \{1, 3, 5, 10\}$.}
\label{tab:result_overall}
%
%

\begin{subtable}[h]{0.49\textwidth}
\captionsetup{font=scriptsize}
\resizebox{\linewidth}{!}{
\begin{tabular}[t]{c@{\hskip 6pt}c@{\hskip 6pt}cccc}
\toprule
Method & Pre-training & 1-shot & 3-shot & 5-shot & 10-shot  \cr \smidrule
TFA~\cite{wang2020frustratingly} & \cellcolor{White} &
23.4 \scriptsize{$\pm$ 4.6} & 29.2 \scriptsize{$\pm$ 1.8} & 32.0 \scriptsize{$\pm$ 1.3} & 35.2 \scriptsize{$\pm$ 1.2}  \cr
\rowcolor{Gray} FSCE-base~\cite{sun2021fsce} & \cellcolor{White}  &
30.5 \scriptsize{$\pm$ 5.3} & 39.4 \scriptsize{$\pm$ 1.6} & 43.9 \scriptsize{$\pm$ 1.1} & 50.2 \scriptsize{$\pm$ 1.1}  \cr
FSCE-con~\cite{sun2021fsce} & \cellcolor{White} &
29.4 \scriptsize{$\pm$ 2.5} & 38.8 \scriptsize{$\pm$ 1.5} & 43.6 \scriptsize{$\pm$ 1.2} & 50.4 \scriptsize{$\pm$ 1.0}  \cr
\rowcolor{Gray} DeFRCN~\cite{qiao2021defrcn} & \cellcolor{White} &
29.3 \scriptsize{$\pm$ 4.2} & 37.8 \scriptsize{$\pm$ 3.1} & 41.6 \scriptsize{$\pm$ 1.9} & 48.2 \scriptsize{$\pm$ 1.9}  \cr
\cmidrule(lr){1-1}\cmidrule(lr){3-6}
Ours-FT & \cellcolor{White} &
\textbf{31.5} \scriptsize{$\pm$ 2.0} & 41.1 \scriptsize{$\pm$ 1.8} & 46.1 \scriptsize{$\pm$ 1.4} & 52.6 \scriptsize{$\pm$ 1.8}  \cr
\rowcolor{Gray} Ours-FSCE+ & \cellcolor{White} \multirow{-6}{*}{COCO} &
31.2 \scriptsize{$\pm$ 2.4} & \textbf{41.3} \scriptsize{$\pm$ 1.5} & \textbf{46.4} \scriptsize{$\pm$ 1.1} & \textbf{53.2} \scriptsize{$\pm$ 1.5}  \cr
\cmidrule(lr){1-1}\cmidrule(lr){2-2}\cmidrule(lr){3-6}
Ours-FT+ & COCO &
35.4 \scriptsize{$\pm$ 1.8} & 44.7 \scriptsize{$\pm$ 1.6} & 49.9 \scriptsize{$\pm$ 1.2} & 56.4 \scriptsize{$\pm$ 1.1}  \cr
\rowcolor{Gray} Ours-FT++ & LVIS++ &
\textbf{35.8} \scriptsize{$\pm$ 3.4} & \textbf{47.2} \scriptsize{$\pm$ 2.1} & \textbf{52.6} \scriptsize{$\pm$ 1.4} & \textbf{59.4} \scriptsize{$\pm$ 1.1}  \cr
\bottomrule
\end{tabular}
}
\caption{Performance with the natural $K$-shot.}
\label{tab:sota_natural}
\end{subtable}
\hspace{\fill}
\begin{subtable}[h]{0.49\textwidth}
\captionsetup{font=scriptsize}
\resizebox{\linewidth}{!}{
\begin{tabular}[t]{cccccc}
\toprule
Method & Pre-training & 1-shot & 3-shot & 5-shot & 10-shot  \cr \smidrule
TFA~\cite{wang2020frustratingly} & \cellcolor{White} &
22.9 \scriptsize{$\pm$ 5.4} & 27.6 \scriptsize{$\pm$ 2.1} & 28.3 \scriptsize{$\pm$ 5.9} & 31.6 \scriptsize{$\pm$ 2.7}  \cr
\rowcolor{Gray} FSCE-con~\cite{sun2021fsce} & \cellcolor{White} &
26.8 \scriptsize{$\pm$ 3.9} & 33.6 \scriptsize{$\pm$ 4.9} & 36.3 \scriptsize{$\pm$ 5.0} & 40.2 \scriptsize{$\pm$ 5.4}  \cr
DeFRCN~\cite{qiao2021defrcn} & \cellcolor{White} &
27.5 \scriptsize{$\pm$ 5.1} & 34.8 \scriptsize{$\pm$ 2.2} & 37.4 \scriptsize{$\pm$ 1.7} & 40.1 \scriptsize{$\pm$ 6.5}  \cr
\cmidrule(lr){1-1}\cmidrule(lr){3-6}
\rowcolor{Gray} Ours-FT & \cellcolor{White} &
\textbf{28.8} \scriptsize{$\pm$ 2.3} & 34.8 \scriptsize{$\pm$ 3.1} & 37.4 \scriptsize{$\pm$ 4.4} & 42.1 \scriptsize{$\pm$ 5.4}  \cr
Ours-FSCE+ & \multirow{-5}{*}{COCO} &
28.7 \scriptsize{$\pm$ 3.8} & \textbf{36.5} \scriptsize{$\pm$ 5.3} & \textbf{38.2} \scriptsize{$\pm$ 5.0} & \textbf{42.7} \scriptsize{$\pm$ 5.7}  \cr
\bottomrule
\end{tabular}
}
\caption{Performance with the balanced $K$-shot.}
\label{tab:sota_balanced}
\end{subtable}
\hspace{\fill}
\begin{subtable}[h]{0.49\textwidth}
\captionsetup{font=scriptsize}
\resizebox{\linewidth}{!}{
\begin{tabular}[t]{c@{\hskip 6pt}c@{\hskip 6pt}cccc}
\toprule
Architecture & Pre-training & 1-shot & 3-shot & 5-shot & 10-shot  \cr \smidrule
Faster R-CNN & \cellcolor{White} &
31.5 \scriptsize{$\pm$ 2.0} & 41.1 \scriptsize{$\pm$ 1.8} & 46.1 \scriptsize{$\pm$ 1.4} & 52.6 \scriptsize{$\pm$ 1.8}  \cr
\rowcolor{Gray} Cascade R-CNN & \cellcolor{White}  &
31.5 \scriptsize{$\pm$ 2.1} & 41.2 \scriptsize{$\pm$ 1.5} & 45.6 \scriptsize{$\pm$ 1.4} & 52.7 \scriptsize{$\pm$ 1.2}  \cr
CenterNet2 & \cellcolor{White} &
29.1 \scriptsize{$\pm$ 5.2} & 40.2 \scriptsize{$\pm$ 1.9} & 45.3 \scriptsize{$\pm$ 1.8} & 52.5 \scriptsize{$\pm$ 1.9}  \cr
\rowcolor{Gray} RetinaNet & \cellcolor{White} &
25.4 \scriptsize{$\pm$ 4.0} & 34.8 \scriptsize{$\pm$ 2.2} & 40.6 \scriptsize{$\pm$ 1.7} & 48.8 \scriptsize{$\pm$ 1.2}  \cr
Deformable-DETR& \cellcolor{White} &
32.0 \scriptsize{$\pm$ 2.9} & 42.3 \scriptsize{$\pm$ 1.6} & 47.4 \scriptsize{$\pm$ 1.4} & 54.7 \scriptsize{$\pm$ 1.0}  \cr
\rowcolor{Gray} Cascade R-CNN-P67 & \cellcolor{White} \multirow{-6}{*}{COCO} &
\textbf{35.4} \scriptsize{$\pm$ 1.8} & \textbf{44.7} \scriptsize{$\pm$ 1.6} & \textbf{49.9} \scriptsize{$\pm$ 1.2} & \textbf{56.4} \scriptsize{$\pm$ 1.1}  \cr
\cmidrule(lr){1-1}\cmidrule(lr){2-2}\cmidrule(lr){3-6}
Faster R-CNN & \cellcolor{Gray} &
31.7 \scriptsize{$\pm$ 1.7} & 41.6 \scriptsize{$\pm$ 1.5} & 46.4 \scriptsize{$\pm$ 1.2} & 53.6 \scriptsize{$\pm$ 1.0}  \cr
\rowcolor{Gray} CenterNet2 & \cellcolor{Gray} &
28.1 \scriptsize{$\pm$ 3.3} & 39.0 \scriptsize{$\pm$ 1.7} & 44.3 \scriptsize{$\pm$ 1.2} & 51.6 \scriptsize{$\pm$ 1.0}  \cr
Cascade R-CNN-P67 & \cellcolor{Gray} \multirow{-3}{*}{LVIS} &
\textbf{34.4} \scriptsize{$\pm$ 1.2} & \textbf{44.0} \scriptsize{$\pm$ 1.6} & \textbf{48.7} \scriptsize{$\pm$ 1.4} & \textbf{55.6} \scriptsize{$\pm$ 1.0}  \cr
\bottomrule
\end{tabular}
}
\caption{The effect of different architectures.}
\label{tab:arch_overall}
\end{subtable}
\hspace{\fill}
\begin{subtable}[h]{0.49\textwidth}
\captionsetup{font=scriptsize}
\resizebox{\linewidth}{!}{
\begin{tabular}[t]{c@{\hskip 6pt}ccccc}
\toprule
Architecture & Pre-training & 1-shot & 3-shot & 5-shot & 10-shot  \cr \smidrule
 & ImageNet &
13.5 \scriptsize{$\pm$ 1.6} & 23.2 \scriptsize{$\pm$ 1.5} & 29.5 \scriptsize{$\pm$ 1.2} & 37.7 \scriptsize{$\pm$ 1.2}  \cr
\rowcolor{Gray} \cellcolor{White} & COCO &
\textbf{35.4} \scriptsize{$\pm$ 1.8} & 44.7 \scriptsize{$\pm$ 1.6} & 49.9 \scriptsize{$\pm$ 1.2} & 56.4 \scriptsize{$\pm$ 1.1}  \cr
& FSODD &
26.7 \scriptsize{$\pm$ 2.9} & 36.9 \scriptsize{$\pm$ 1.5} & 42.3 \scriptsize{$\pm$ 1.2} & 49.1 \scriptsize{$\pm$ 1.0}  \cr
\rowcolor{Gray} \cellcolor{White} & LVIS &
34.4 \scriptsize{$\pm$ 2.0} & 44.0 \scriptsize{$\pm$ 1.6} & 48.7 \scriptsize{$\pm$ 1.4} & 55.6 \scriptsize{$\pm$ 1.0}  \cr
& Unified &
33.3 \scriptsize{$\pm$ 2.2} & 44.3 \scriptsize{$\pm$ 1.4} & 49.6 \scriptsize{$\pm$ 1.2} & 56.6 \scriptsize{$\pm$ 1.1}  \cr
\rowcolor{Gray} \cellcolor{White} \multirow{-6}{*}{Cascade R-CNN-P67~} & LVIS+ &
34.7 \scriptsize{$\pm$ 4.2} & \textbf{46.6} \scriptsize{$\pm$ 1.6} & \textbf{52.0} \scriptsize{$\pm$ 1.0} & \textbf{59.0} \scriptsize{$\pm$ 1.0}  \cr
\cmidrule(lr){1-1}\cmidrule(lr){2-2}\cmidrule(lr){3-6}
\cellcolor{Gray} & COCO &
29.1 \scriptsize{$\pm$ 5.2} & 40.2 \scriptsize{$\pm$ 1.9} & 45.3 \scriptsize{$\pm$ 1.8} & 52.5 \scriptsize{$\pm$ 1.9}  \cr
\rowcolor{Gray} & LVIS &
28.1 \scriptsize{$\pm$ 3.3} & 39.0 \scriptsize{$\pm$ 1.7} & 44.3 \scriptsize{$\pm$ 1.2} & 51.6 \scriptsize{$\pm$ 1.0}  \cr
\cellcolor{Gray} & LVIS+ &
34.9 \scriptsize{$\pm$ 3.2} & 46.5 \scriptsize{$\pm$ 1.9} & 51.8 \scriptsize{$\pm$ 1.3} & 58.8 \scriptsize{$\pm$ 1.0}  \cr
\rowcolor{Gray} \multirow{-4}{*}{CenterNet2} & LVIS++ &
\textbf{35.8} \scriptsize{$\pm$ 3.4} & \textbf{47.2} \scriptsize{$\pm$ 2.1} & \textbf{52.6} \scriptsize{$\pm$ 1.4} & \textbf{59.4} \scriptsize{$\pm$ 1.1}  \cr
\bottomrule
\end{tabular}
}
\caption{The effect of different pre-training.}
\label{tab:pre_training_overall}
\end{subtable}
\end{table*}

%% file: table_arch_5shot.tex
\begin{table*}[t]
\def\smidrulea{\cmidrule(rl){1-2}\cmidrule(rl){3-3}\cmidrule(rl){4-4}\cmidrule(rl){5-5}\cmidrule(rl){6-6}\cmidrule(rl){7-7}\cmidrule(rl){8-8}\cmidrule(rl){9-9}\cmidrule(rl){10-10}\cmidrule(rl){11-11}\cmidrule(rl){12-12}}
\def\smidruleb{\cmidrule(rl){1-1}\cmidrule(rl){2-2}\cmidrule(rl){3-3}\cmidrule(rl){4-4}\cmidrule(rl){5-5}\cmidrule(rl){6-6}\cmidrule(rl){7-7}\cmidrule(rl){8-8}\cmidrule(rl){9-9}\cmidrule(rl){10-10}\cmidrule(rl){11-11}\cmidrule(rl){12-12}\cmidrule(rl){13-13}\cmidrule(rl){14-14}}
\def\smidrulec{\cmidrule(rl){1-1}\cmidrule(rl){2-2}\cmidrule(rl){3-3}\cmidrule(rl){4-4}\cmidrule(rl){5-5}\cmidrule(rl){6-6}\cmidrule(rl){7-7}\cmidrule(rl){8-8}\cmidrule(rl){9-9}\cmidrule(rl){10-10}\cmidrule(rl){11-11}}
\def\smidruled{\cmidrule(rl){1-1}\cmidrule(rl){2-2}\cmidrule(rl){3-3}\cmidrule(rl){4-4}\cmidrule(rl){5-5}\cmidrule(rl){6-6}\cmidrule(rl){7-7}\cmidrule(rl){8-8}\cmidrule(rl){9-9}\cmidrule(rl){10-10}\cmidrule(rl){11-11}\cmidrule(rl){12-12}\cmidrule(rl){13-13}}
\centering
\caption{Per-dataset 5-shot performance of the effects of tuning different parameters, different architectures and pre-training datasets.}
\label{tab:5_shot}
\begin{subtable}[h]{\textwidth}
\captionsetup{font=scriptsize}
\resizebox{\linewidth}{!}{
\begin{tabular}{ccccccccccccc}
\toprule
5-shot & Aerial & Agriculture & Animal & Cartoon & Fashion & Food & Logo & Person & Security & Traffic & \multirow{2}{*}{Mean} & \multirow{2}{*}{Rank} \cr \smidrulec
Unfrozen & VisDrone & DeepFruits & iWildCam & Clipart & iMaterialist & Oktoberfest & LogoDet-3K & CrowdHuman & SIXray & KITTI \cr \smidruled
Last FC layers (TFA~\cite{wang2020frustratingly}) & 
10.1 \scriptsize{$\pm$ 0.6} & 47.5 \scriptsize{$\pm$ 1.8} & \textbf{71.7} \scriptsize{$\pm$ 2.4} & 40.2 \scriptsize{$\pm$ 2.6} & 8.0 \scriptsize{$\pm$ 0.9} & 41.1 \scriptsize{$\pm$ 4.8} & 14.2 \scriptsize{$\pm$ 3.7} & 30.6 \scriptsize{$\pm$ 0.8} & 7.9 \scriptsize{$\pm$ 2.0} & \textbf{48.3} \scriptsize{$\pm$ 3.4} & 32.0 \scriptsize{$\pm$ 1.3} & 2.7 \scriptsize{$\pm$ 0.2} \cr
\rowcolor{Gray} Detection head (FSCE-base~\cite{sun2021fsce}) & 
13.0 \scriptsize{$\pm$ 0.7} & 59.2 \scriptsize{$\pm$ 3.2} & 70.6 \scriptsize{$\pm$ 1.7} & 43.5 \scriptsize{$\pm$ 2.5} & 20.5 \scriptsize{$\pm$ 0.9} & 70.7 \scriptsize{$\pm$ 3.5} & 47.2 \scriptsize{$\pm$ 3.5} &\textbf{51.6} \scriptsize{$\pm$ 2.0} & 15.9 \scriptsize{$\pm$ 2.2} & 47.1 \scriptsize{$\pm$ 4.3} & 43.9 \scriptsize{$\pm$ 1.1} & 1.9 \scriptsize{$\pm$ 0.2} \cr
Whole network (Ours-FT) & 
\textbf{14.2} \scriptsize{$\pm$ 0.8} & \textbf{60.7} \scriptsize{$\pm$ 3.8} & 63.3 \scriptsize{$\pm$ 5.7} & \textbf{49.3} \scriptsize{$\pm$ 3.5} & \textbf{21.3} \scriptsize{$\pm$ 0.8} & \textbf{81.6} \scriptsize{$\pm$ 4.0} & \textbf{49.8} \scriptsize{$\pm$ 3.5} & 51.5 \scriptsize{$\pm$ 2.2} & \textbf{23.9} \scriptsize{$\pm$ 3.9} & 45.3 \scriptsize{$\pm$ 3.7} & \textbf{46.1} \scriptsize{$\pm$ 1.4} & \textbf{1.5} \scriptsize{$\pm$ 0.2} \cr
\bottomrule
\end{tabular}
}
\caption{\scriptsize{Fine-tuning different number of parameters with Faster R-CNN pre-trained on COCO.
}}
\label{tab:tune}
\end{subtable}

\begin{subtable}[h]{\textwidth}
\resizebox{\linewidth}{!}{
\begin{tabular}{c@{\hskip 6pt}c@{\hskip 6pt}cccccccccccc}
\toprule
\multicolumn{2}{c}{5-shot} & Aerial & Agriculture & Animal & Cartoon & Fashion & Food & Logo & Person & Security & Traffic & \multirow{2}{*}{Mean} & \multirow{2}{*}{Rank} \cr \smidrulea
Architecture & Pre-training & VisDrone & DeepFruits & iWildCam & Clipart & iMaterialist & Oktoberfest & LogoDet-3K & CrowdHuman & SIXray & KITTI \cr \smidruleb
Faster R-CNN & \cellcolor{White} & 14.2 \scriptsize{$\pm$ 0.8} & 60.7 \scriptsize{$\pm$ 3.8} & 63.3 \scriptsize{$\pm$ 5.7} & 49.3 \scriptsize{$\pm$ 3.5} & 21.3 \scriptsize{$\pm$ 0.8} & 81.6 \scriptsize{$\pm$ 4.0} & 49.8 \scriptsize{$\pm$ 3.5} & 51.5 \scriptsize{$\pm$ 2.2} & 23.9 \scriptsize{$\pm$ 3.9} & 45.3 \scriptsize{$\pm$ 3.7} & 46.1 \scriptsize{$\pm$ 1.4} & 3.4 \scriptsize{$\pm$ 0.3} \cr
\rowcolor{Gray} Cascade R-CNN & \cellcolor{White} & 13.0 \scriptsize{$\pm$ 0.8} & 58.9 \scriptsize{$\pm$ 3.3} & 66.8 \scriptsize{$\pm$ 3.7} & 50.8 \scriptsize{$\pm$ 2.8} & 20.5 \scriptsize{$\pm$ 0.6} & 80.5 \scriptsize{$\pm$ 2.2} & 48.5 \scriptsize{$\pm$ 4.7} & 51.4 \scriptsize{$\pm$ 2.0} & 21.6 \scriptsize{$\pm$ 4.4} & 44.4 \scriptsize{$\pm$ 4.1} & 45.6 \scriptsize{$\pm$ 1.4} & 4.1 \scriptsize{$\pm$ 0.4} \cr
CenterNet2 & \cellcolor{White} &13.5 \scriptsize{$\pm$ 0.7} & 59.0 \scriptsize{$\pm$ 4.7} & 61.6 \scriptsize{$\pm$ 4.7} & 49.2 \scriptsize{$\pm$ 7.6} & 22.2 \scriptsize{$\pm$ 1.9} & 79.7 \scriptsize{$\pm$ 3.5} & 51.0 \scriptsize{$\pm$ 4.2} & 51.4 \scriptsize{$\pm$ 3.7} & 22.2 \scriptsize{$\pm$ 4.8} & 43.7 \scriptsize{$\pm$ 4.9} & 45.3 \scriptsize{$\pm$ 1.8} & 3.9 \scriptsize{$\pm$ 0.3} \cr
\rowcolor{Gray} RetinaNet & \cellcolor{White} & 9.9 \scriptsize{$\pm$ 0.6} & 55.8 \scriptsize{$\pm$ 2.7} & 59.2 \scriptsize{$\pm$ 6.8} & 26.8 \scriptsize{$\pm$ 2.2} & 17.6 \scriptsize{$\pm$ 0.4} & 79.5 \scriptsize{$\pm$ 4.0} & 49.2 \scriptsize{$\pm$ 3.5} & 47.6 \scriptsize{$\pm$ 1.9} & 20.6 \scriptsize{$\pm$ 3.3} & 39.7 \scriptsize{$\pm$ 3.3} & 40.6 \scriptsize{$\pm$ 1.7} & 5.4 \scriptsize{$\pm$ 0.5} \cr
Deformable-DETR & \cellcolor{White} & 15.0 \scriptsize{$\pm$ 0.6} & \textbf{68.8} \scriptsize{$\pm$ 4.3} & 66.0 \scriptsize{$\pm$ 4.3} & 43.7 \scriptsize{$\pm$ 1.6} & 22.4 \scriptsize{$\pm$ 1.1} & 77.2 \scriptsize{$\pm$ 4.6} & 50.3 \scriptsize{$\pm$ 3.5} & \textbf{56.7} \scriptsize{$\pm$ 2.4} & 26.3 \scriptsize{$\pm$ 3.8} & \textbf{47.9} \scriptsize{$\pm$ 3.2} & 47.4 \scriptsize{$\pm$ 1.4} & 2.6 \scriptsize{$\pm$ 0.5} \cr
\rowcolor{Gray} Cascade R-CNN-P67 & \cellcolor{White} \multirow{-6}{*}{COCO} & \textbf{15.9} \scriptsize{$\pm$ 0.8} & 63.1 \scriptsize{$\pm$ 2.3} & \textbf{73.7} \scriptsize{$\pm$ 2.8} & \textbf{56.6} \scriptsize{$\pm$ 2.3} & \textbf{24.2} \scriptsize{$\pm$ 0.8} & \textbf{83.2} \scriptsize{$\pm$ 2.4} & \textbf{54.5} \scriptsize{$\pm$ 4.4} & 53.6 \scriptsize{$\pm$ 1.8} & \textbf{26.4} \scriptsize{$\pm$ 3.9} & 47.6 \scriptsize{$\pm$ 3.6} & \textbf{49.9} \scriptsize{$\pm$ 1.2} & \textbf{1.5} \scriptsize{$\pm$ 0.3} \cr
\cmidrule(lr){1-1}\cmidrule(lr){2-2}\cmidrule(lr){3-14}
Faster R-CNN & \cellcolor{Gray} & 14.2 \scriptsize{$\pm$ 0.7} & \textbf{66.2} \scriptsize{$\pm$ 3.4} & 69.3 \scriptsize{$\pm$ 3.7} & 39.8 \scriptsize{$\pm$ 1.8} & 28.0 \scriptsize{$\pm$ 0.6} & 80.7 \scriptsize{$\pm$ 2.7} & 51.3 \scriptsize{$\pm$ 4.2} & 48.3 \scriptsize{$\pm$ 2.2} & 24.7 \scriptsize{$\pm$ 3.6} & 41.3 \scriptsize{$\pm$ 3.5} & 46.4 \scriptsize{$\pm$ 1.2} & 2.1 \scriptsize{$\pm$ 0.3} \cr
\rowcolor{Gray} CenterNet2 & \cellcolor{Gray} & 13.3 \scriptsize{$\pm$ 0.7} & 64.6 \scriptsize{$\pm$ 3.5} & 50.2 \scriptsize{$\pm$ 25.3} & 34.1 \scriptsize{$\pm$ 2.5} & 25.6 \scriptsize{$\pm$ 0.5} & 81.6 \scriptsize{$\pm$ 2.8} & 53.7 \scriptsize{$\pm$ 3.8} & 46.6 \scriptsize{$\pm$ 2.0} & 22.8 \scriptsize{$\pm$ 3.3} & 38.9 \scriptsize{$\pm$ 3.9} & 43.1 \scriptsize{$\pm$ 6.9} & 2.7 \scriptsize{$\pm$ 0.3} \cr
Cascade R-CNN-P67 & \cellcolor{Gray} \multirow{-3}{*}{LVIS} & \textbf{15.2} \scriptsize{$\pm$ 0.7} & 65.4 \scriptsize{$\pm$ 2.0} & \textbf{71.7} \scriptsize{$\pm$ 2.0} & \textbf{46.0} \scriptsize{$\pm$ 2.5} & \textbf{30.2} \scriptsize{$\pm$ 0.6} & \textbf{81.9} \scriptsize{$\pm$ 3.0} & \textbf{56.6} \scriptsize{$\pm$ 5.0} & \textbf{49.6} \scriptsize{$\pm$ 1.7} & 25.9 \scriptsize{$\pm$ 4.3} & \textbf{44.7} \scriptsize{$\pm$ 3.7} & \textbf{48.7} \scriptsize{$\pm$ 1.4} & \textbf{1.2} \scriptsize{$\pm$ 0.3} \cr
\bottomrule
\end{tabular}
}
\caption{\scriptsize{Performance of different architectures pre-trained on COCO and LVIS.}}
\label{tab:arch}
\end{subtable}

\begin{subtable}[h]{\textwidth}
\resizebox{\linewidth}{!}{
\begin{tabular}{c@{\hskip 6pt}ccccccccccccc}
\toprule
\multicolumn{2}{c}{5-shot} & Aerial & Agriculture & Animal & Cartoon & Fashion & Food & Logo & Person & Security & Traffic & \multirow{2}{*}{Mean} & \multirow{2}{*}{Rank} \cr \smidrulea
Architecture & Pre-training & VisDrone & DeepFruits & iWildCam & Clipart & iMaterialist & Oktoberfest & LogoDet-3K & CrowdHuman & SIXray & KITTI \cr \smidruleb
 & ImageNet & 9.8 \scriptsize{$\pm$ 0.5} & 53.0 \scriptsize{$\pm$ 3.5} & 7.4 \scriptsize{$\pm$ 3.2} & 13.1 \scriptsize{$\pm$ 2.5} & 19.2 \scriptsize{$\pm$ 0.6} & 77.4 \scriptsize{$\pm$ 3.4} & 46.4 \scriptsize{$\pm$ 4.4} & 32.0 \scriptsize{$\pm$ 3.1} & 13.1 \scriptsize{$\pm$ 3.2} & 23.8 \scriptsize{$\pm$ 3.4} & 29.5 \scriptsize{$\pm$ 1.2} & 6.0 \scriptsize{$\pm$ 0.0} \cr
\rowcolor{Gray} \cellcolor{White} & COCO & 15.9 \scriptsize{$\pm$ 0.8} & 63.1 \scriptsize{$\pm$ 2.3} & \textbf{73.7} \scriptsize{$\pm$ 2.8} & \textbf{56.6} \scriptsize{$\pm$ 2.3} & 24.2 \scriptsize{$\pm$ 0.8} & 83.2 \scriptsize{$\pm$ 2.4} & 54.5 \scriptsize{$\pm$ 4.4} & \textbf{53.6} \scriptsize{$\pm$ 1.8} & 26.4 \scriptsize{$\pm$ 3.9} & 47.6 \scriptsize{$\pm$ 3.6} & 49.9 \scriptsize{$\pm$ 1.2} & 2.8 \scriptsize{$\pm$ 0.4} \cr
 & FSODD & 10.6 \scriptsize{$\pm$ 0.6} & 67.5 \scriptsize{$\pm$ 3.7} & 64.5 \scriptsize{$\pm$ 3.4} & 26.5 \scriptsize{$\pm$ 2.0} & 21.2 \scriptsize{$\pm$ 0.5} & 82.9 \scriptsize{$\pm$ 2.9} & 55.7 \scriptsize{$\pm$ 3.8} & 38.4 \scriptsize{$\pm$ 2.2} & 23.2 \scriptsize{$\pm$ 3.7} & 32.3 \scriptsize{$\pm$ 3.6} & 42.3 \scriptsize{$\pm$ 1.2} & 4.3 \scriptsize{$\pm$ 0.4} \cr
\rowcolor{Gray} \cellcolor{White} & LVIS & 15.2 \scriptsize{$\pm$ 0.7} & 65.4 \scriptsize{$\pm$ 2.0} & 71.7 \scriptsize{$\pm$ 2.0} & 46.0 \scriptsize{$\pm$ 2.5} & 30.2 \scriptsize{$\pm$ 0.6} & 81.9 \scriptsize{$\pm$ 3.0} & 56.6 \scriptsize{$\pm$ 5.0} & 49.6 \scriptsize{$\pm$ 1.7} & 25.9 \scriptsize{$\pm$ 4.3} & 44.7 \scriptsize{$\pm$ 3.7} & 48.7 \scriptsize{$\pm$ 1.4} & 3.4 \scriptsize{$\pm$ 0.4} \cr
 & Unified & 16.3 \scriptsize{$\pm$ 0.9} & 68.8 \scriptsize{$\pm$ 3.0} & 66.6 \scriptsize{$\pm$ 4.0} & 48.5 \scriptsize{$\pm$ 2.9} & 26.4 \scriptsize{$\pm$ 0.7} & \textbf{83.6} \scriptsize{$\pm$ 2.9} & 59.3 \scriptsize{$\pm$ 3.7} & 52.9 \scriptsize{$\pm$ 1.6} & 27.2 \scriptsize{$\pm$ 3.6} & 46.1 \scriptsize{$\pm$ 3.9} & 49.6 \scriptsize{$\pm$ 1.2} & 2.5 \scriptsize{$\pm$ 0.4} \cr
\rowcolor{Gray} \cellcolor{White} \multirow{-6}{*}{Cascade R-CNN-P67~} & LVIS+ & \textbf{18.5} \scriptsize{$\pm$ 1.0} & \textbf{76.8} \scriptsize{$\pm$ 3.3} & 59.5 \scriptsize{$\pm$ 4.1} & 52.6 \scriptsize{$\pm$ 2.0} & \textbf{31.5} \scriptsize{$\pm$ 0.7} & 82.5 \scriptsize{$\pm$ 3.3} & \textbf{61.2} \scriptsize{$\pm$ 2.7} & 52.3 \scriptsize{$\pm$ 1.9} & \textbf{36.0} \scriptsize{$\pm$ 2.5} & \textbf{49.6} \scriptsize{$\pm$ 3.9} & \textbf{52.0} \scriptsize{$\pm$ 1.1} & \textbf{1.9} \scriptsize{$\pm$ 0.5} \cr
\cmidrule(lr){1-1}\cmidrule(lr){2-2}\cmidrule(lr){3-14}
\cellcolor{Gray} & COCO & 13.5 \scriptsize{$\pm$ 0.7} & 59.0 \scriptsize{$\pm$ 4.7} & 61.6 \scriptsize{$\pm$ 4.7} & 49.2 \scriptsize{$\pm$ 7.6} & 22.2 \scriptsize{$\pm$ 1.9} & 79.7 \scriptsize{$\pm$ 3.5} & 51.0 \scriptsize{$\pm$ 4.2} & 51.4 \scriptsize{$\pm$ 3.7} & 22.2 \scriptsize{$\pm$ 4.8} & 43.7 \scriptsize{$\pm$ 4.9} & 45.3 \scriptsize{$\pm$ 1.8} & 3.2 \scriptsize{$\pm$ 0.3} \cr
\rowcolor{Gray} & LVIS & 13.3 \scriptsize{$\pm$ 0.7} & 64.6 \scriptsize{$\pm$ 3.5} & 61.5 \scriptsize{$\pm$ 4.3} & 34.1 \scriptsize{$\pm$ 2.5} & 25.6 \scriptsize{$\pm$ 0.5} & \textbf{81.6} \scriptsize{$\pm$ 2.8} & 53.7 \scriptsize{$\pm$ 3.8} & 46.6 \scriptsize{$\pm$ 2.0} & 22.8 \scriptsize{$\pm$ 3.3} & 38.9 \scriptsize{$\pm$ 3.9} & 44.3 \scriptsize{$\pm$ 1.2} & 3.2 \scriptsize{$\pm$ 0.2} \cr
\cellcolor{Gray} & LVIS+ & \textbf{18.2} \scriptsize{$\pm$ 0.9} & 74.0 \scriptsize{$\pm$ 3.3} & 63.7 \scriptsize{$\pm$ 3.8} & 50.5 \scriptsize{$\pm$ 1.8} & 31.5 \scriptsize{$\pm$ 0.8} & 80.4 \scriptsize{$\pm$ 4.0} & \textbf{62.3} \scriptsize{$\pm$ 3.2} & \textbf{54.0} \scriptsize{$\pm$ 2.1} & 37.3 \scriptsize{$\pm$ 4.1} & 46.5 \scriptsize{$\pm$ 4.7} & 51.8 \scriptsize{$\pm$ 1.3} & 2.0 \scriptsize{$\pm$ 0.4} \cr
\rowcolor{Gray} \multirow{-4}{*}{CenterNet2} & LVIS++ & 18.1 \scriptsize{$\pm$ 0.9} & \textbf{77.5} \scriptsize{$\pm$ 2.4} & \textbf{64.4} \scriptsize{$\pm$ 4.7} & \textbf{52.1} \scriptsize{$\pm$ 1.8} & \textbf{33.1} \scriptsize{$\pm$ 0.7} & 80.4 \scriptsize{$\pm$ 3.5} & 61.0 \scriptsize{$\pm$ 4.5} & 54.7 \scriptsize{$\pm$ 2.0} & \textbf{37.8} \scriptsize{$\pm$ 3.2} & \textbf{46.9} \scriptsize{$\pm$ 4.1} & \textbf{52.6} \scriptsize{$\pm$ 1.4} & \textbf{1.5} \scriptsize{$\pm$ 0.3} \cr
\bottomrule
\end{tabular}
}
\caption{\scriptsize{Performance of Cascade R-CNN-P67 and CenterNet2 pre-trained on different datasets.}}
\label{tab:pre_training}
\end{subtable}
\end{table*}

%% file: 5_conclusion.tex
\section{Conclusion}
\label{sec:conclusion}

We present the Multi-dOmain Few-Shot Object Detection (MoFSOD) benchmark consisting of 10 datasets from different domains to evaluate FSOD methods. Under the proposed benchmark, we conducted extensive experiments on the impact of freezing parameters, different architectures, and different pre-training datasets. Based on our findings, we proposed simple extensions of the existing methods and achieved state-of-the-art results on the benchmark. In the future, we would like to go beyond empirical studies and modifications, to designing architectures and smart-tuning methods for a wide range of FSOD tasks.

%% file: appendix.tex
\appendix
\numberwithin{table}{section}
\numberwithin{figure}{section}
\numberwithin{equation}{section}
\numberwithin{algorithm}{section}

\section{Natural $K$-Shot Sampling}
\label{sec:natural_k_shot}
In this section, we describe how we perform natural $K$-shot sampling in detail:

\textbf{Step 1. Sample $C \times K$ images.} $C$ is the number of classes of the original dataset $\mathcal{S}$. In this step, without worrying about class labels, we sample $\mathcal{S}$ from the entire dataset $\mathcal{D}$. Unlike the standard $K$-shot sampling algorithm in recent FSOD works~\cite{kang2019few,wang2020frustratingly,sun2021fsce}, we do not apply stratified sampling. This is because an image usually contains multiple annotations, such that stratified sampling might result in an artificial class distribution~\cite{kang2019few}.

\textbf{Step 2. Check missing classes.} The initial sampled dataset $\mathcal {} $ might not contain some classes, particularly those present only in a few images in the original dataset.  To compensate for this, we check if there are any missing classes and update the sampled dataset. Specifically, we manage two datasets: $\mathcal{P}$ is a set of images to be added, and $\mathcal{Q}$ is a set of images to be kept. Then, for each class, if no image in $\mathcal{S}$ contains the class, we sample an image from the $\mathcal{D}$ containing the class and put it in $\mathcal{P}$; otherwise, we sample an image from $\mathcal{S}$ containing the class and put it in $\mathcal{Q}$.

\textbf{Step 3. Update the sampled dataset.} As the final step, we adjust the initial dataset $\mathcal{S}$ to guarantee that all classes are present.
To match the number of added and removed images, we sample a set of images to be removed $\mathcal{R}$ from $\mathcal{S} - \mathcal{Q}$ where the size of $\mathcal{R}$ is the same as $\mathcal{P}$. Here, $\mathcal{Q}$ guarantees that any class in $\mathcal{S}$ does not become empty.
\input{alg_natural}
Finally, we add $\mathcal{P}$ and remove $\mathcal{R}$ from $\mathcal{S}$. 

The complete algorithm is in \Cref{alg:natural}.

\section{Dataset Size Reduction}
We initially collected more than 100 public detection datasets, and then selected 32 datasets based on availability, diversity of domains, annotation quality, and number of citations. After initial experiments on them, to reduce the computational burden for future research, we picked 10 datasets out of the 32, which show similar performance trends with the 32 datasets, while covering a variety of domains based on the domain distance.

In the proposed MoFSOD benchmark, several datasets contain a large number of classes and testing images, such as LogoDet-3K. With the proposed natural $K$-shot sampling, the training time is proportional to the number of classes. To address concerns on computational cost and speed up overall experiment time,  we limited the number of classes to 50 and the number of test samples to 1k.

Specifically, we randomly sample 50 classes and remove images containing all the rest classes in each episode, such that the intention of the original datasets is kept, \ie, all remaining logos or traffic signs should be detected. We note that all classes in these datasets are mostly isolated to certain images, such that removing images containing a class does not hurt the distribution of other classes. We confirmed that the performance differences between sampled and full
test sets are less than 1.5\% for all datasets.

\section{Additional Experimental Results}
\subsection{Dataset Statistics}
More detailed statistics of the ten datasets of MoFSOD can be found in~\Cref{tb:coreset}.
\input{table_coreset}

\input{table_arch_1shot}
\input{table_arch_3shot}
\input{table_arch_10shot}
\subsection{Detailed 1-, 3- and 10-shot Results}
In addition to per-dataset 5-shot results in \tablethree~ of the main paper, we present per-dataset 1-, 3- and 10-shot results in \Cref{tab:1_shot}, \ref{tab:3_shot}, and \ref{tab:10_shot}, respectively.

\subsection{Extended 32 Datasets Results}
\begin{figure*}[t]
\centering
\includegraphics[width=.495\linewidth]{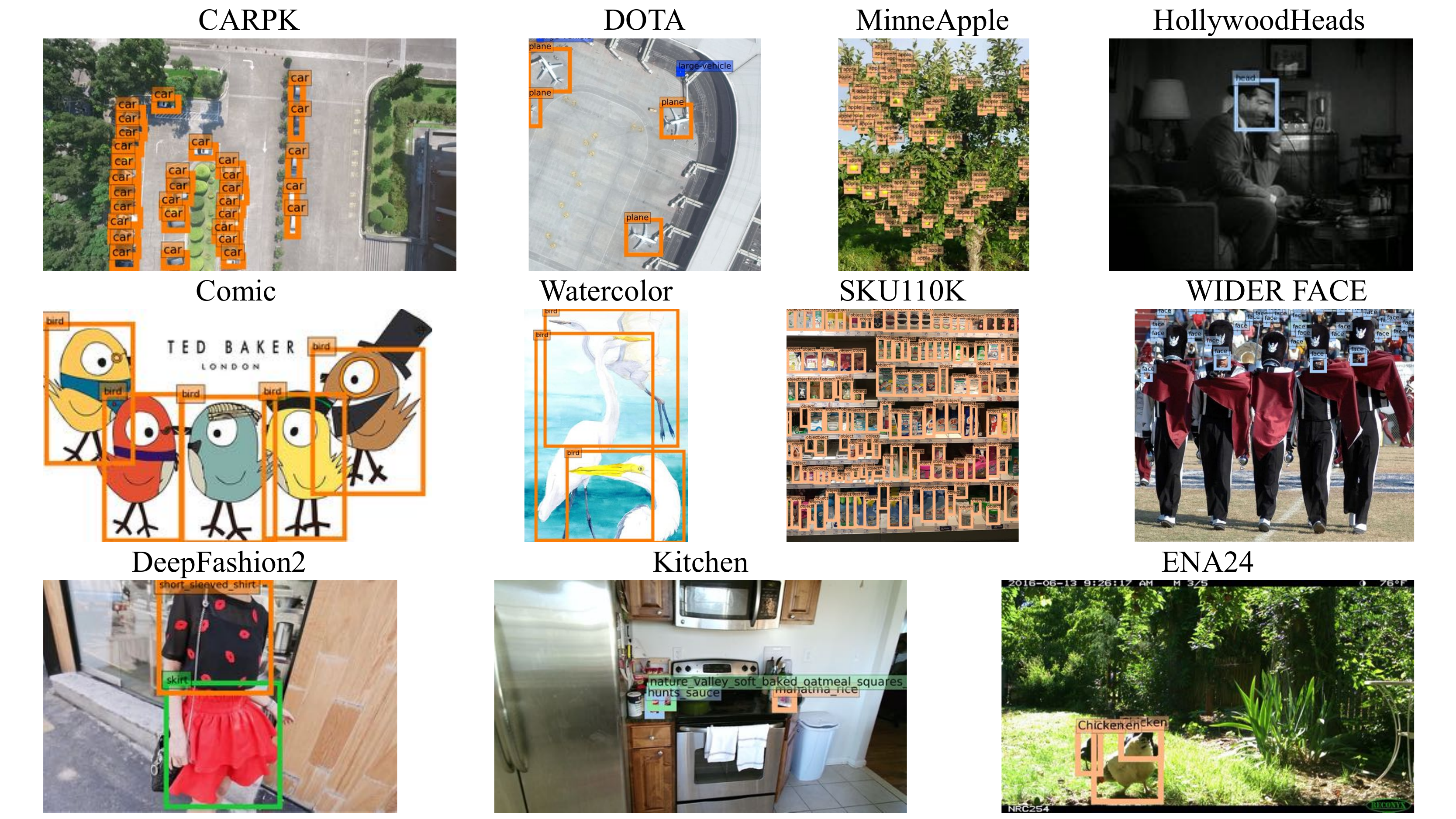}
\includegraphics[width=.495\linewidth]{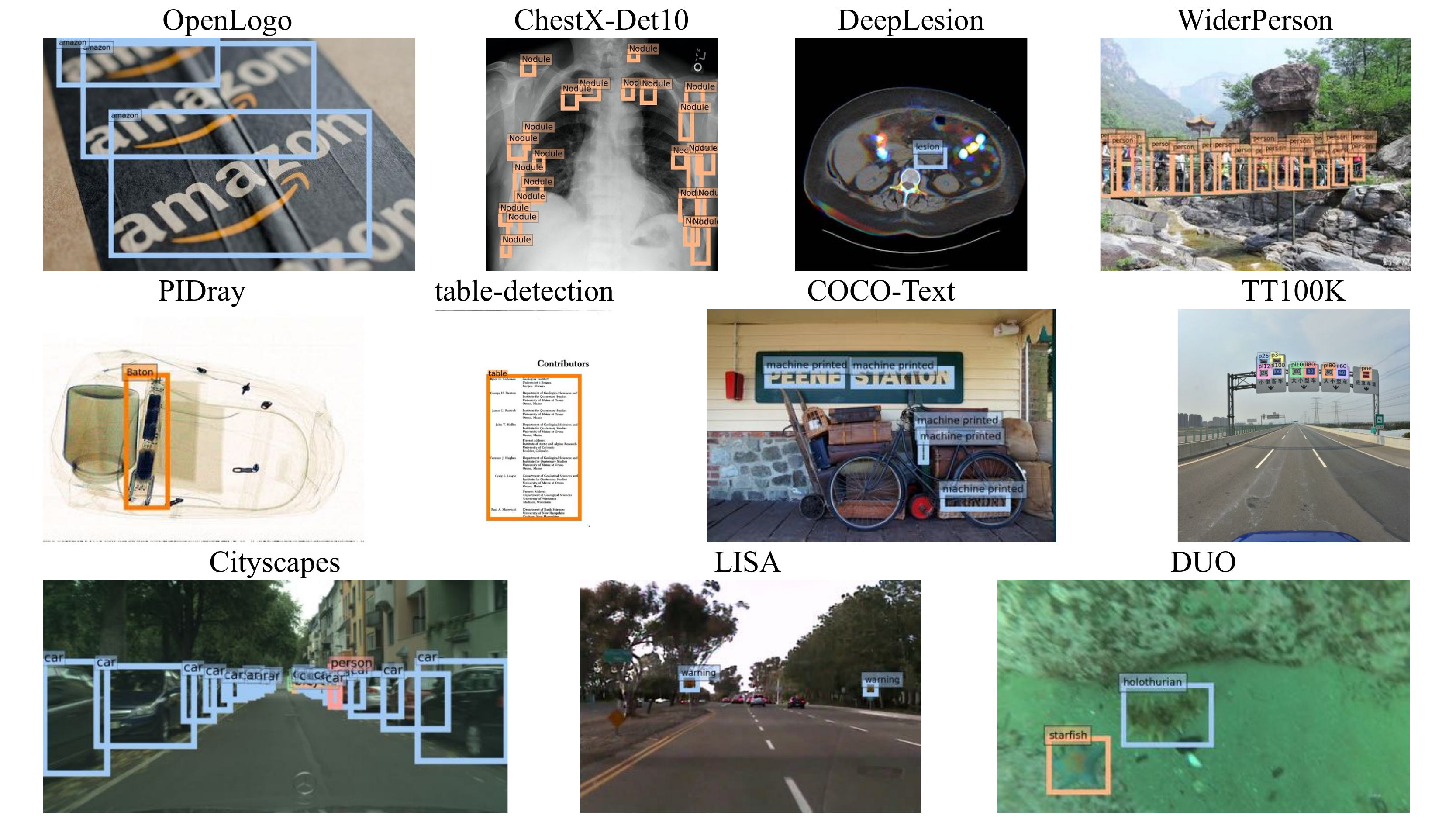}
\caption{Image samples of the additional datasets.}
\label{fig:datasets_12}
\end{figure*}

We evaluate \textit{FT} against SOTA methods in an extended 32 dataset benchmark on 17 Domains. These datasets are: 
CARPK~\cite{carpk}, DOTA~\cite{dota}, and VisDrone~\cite{visdrone} in aerial images,
DeepFruits~\cite{deepfruits} and MinneApple~\cite{minneapple} in agriculture,
ENA24~\cite{ena24} and iWildCam~\cite{iwildcam} in animal in the wild,
Clipart, Comic, and Watercolor~\cite{cartoon} in cartoon,
SKU110K~\cite{sku110k} in dense product,
DeepFashion2~\cite{deepfashion2} and iMaterialist~\cite{imaterialist} in fashion,
WIDER FACE~\cite{widerface} in face,
Kitchen~\cite{kitchen} and Oktoberfest~\cite{oktoberfest} in food,
HollywoodHeads~\cite{hollywoodheads} in head,
LogoDet-3K~\cite{logodet3k} and OpenLogo~\cite{openlogo} in logo,
ChestX-Det10~\cite{chest10} and DeepLesion~\cite{deeplesion} in medical imaging,
CrowdHuman~\cite{crowdhuman} and WiderPerson~\cite{widerperson} in person,
PIDray~\cite{pidray} and SIXray~\cite{sixray} in security,
table-detection~\cite{table} in table,
COCO-Text~\cite{cocotext} in text in the wild, and
Cityscapes~\cite{cityscapes}, KITTI~\cite{kitti}, LISA~\cite{lisa}, and TT100K~\cite{tt100k} in traffic,
DUO~\cite{duo} in underwater. 
\input{table_dataset}
Their statistics can be found in~\Cref{tb:dataset}. Sample images from these datasets can be found in~\Cref{fig:datasets_12}.

\Cref{tab:result_all} compares \textit{FT} and SOTA methods. \textit{TFA-cos} is a variation of \textit{TFA} where the classification head is replaced with the cosine similarity. Note that, while the comparison within tables is fair, the results are NOT directly comparable to the results in the main paper, as they are experimented in different settings. Specifically, the architecture uses deformable convolution v1 while the one in the main paper uses v2, and is trained with a non-standard scheduler. We can observe that \textit{FT} is a strong baseline, outperforming all other methods.


\input{table_all}
\input{table_arch}

We then present the results between different architectures and pre-training datasets in~\Cref{tab:result_arch}.
Although the ablation study here is not as comprehensive as the main paper, we can still see that \textit{Cascade R-CNN-P67} outperforms \textit{Faster R-CNN}. The margin here is smaller mainly due to the less optimal learning rate scheduler we used for \textit{Cascade R-CNN-P67} and possibly the lack of deformable convolution in this model. Once we use the same learning rate scheduler and backbone architecture with deformable convolution v2~\cite{zhu19deconvv2} for both \textit{Cascade R-CNN-P67} and \textit{Faster R-CNN} as in the main paper, the performance gap for different shots actually increases. On the other hand, the comparison between all \textit{Cascade R-CNN-P67} experiments is fair. We can see that over a larger range of domains, \textbf{Unified} provides better results than \textbf{COCO} by a significant margin. However, this performance gap could be due to the non-optimal training of \textbf{COCO}. These suggest that besides the size/quality of the pre-training datasets, how to train for downstream tasks optimally is also an important factor.

%% file: alg_natural.tex
\begin{algorithm}[t]
\caption{Natural $K$-shot sampling algorithm.}
\label{alg:natural}
\begin{algorithmic}[1]
\STATE \textbf{Input:} Dataset $\mathcal{D}$, classes $\mathcal{Y}$, number of classes $C = \lvert \mathcal{Y} \rvert$, average shot number $K$
\STATE \textbf{Output:} Sampled dataset $\mathcal{S}$
\IF{$\lvert \mathcal{D} \rvert \leq C \times K$}
    \STATE $\mathcal{S} \leftarrow \mathcal{D}$
\ELSE
    \STATE // Step 1: sample an initial dataset
    \STATE Sample $\mathcal{S} \subset \mathcal{D}$ where $\lvert \mathcal{S} \rvert = N \times K$
    \STATE // Step 2: check missing classes
    \STATE $\mathcal{P} = \{\}$ // images to be added 
    \STATE $\mathcal{Q} = \{\}$ // images to be kept 
    \FOR{$y \in \mathcal{Y}$}
        \IF{no image in $\mathcal{S}$ contains $y$}
            \STATE Sample $I \in \mathcal{D}$ where $I$ contains $y$
            \STATE $\mathcal{P} \leftarrow \mathcal{P} \cup \{ I \}$
        \ELSE
            \STATE Sample $I \in \mathcal{S}$ where $I$ contains $y$
            \STATE $\mathcal{Q} \leftarrow \mathcal{Q} \cup \{ I \}$
        \ENDIF
    \ENDFOR
    \STATE // Step 3: update the sampled dataset
    \IF{$\lvert \mathcal{P} \rvert > 0$}
        \STATE Sample $\mathcal{R} \subset \mathcal{S} - \mathcal{Q}$ where $\lvert \mathcal{R} \rvert = \lvert \mathcal{P} \rvert$
        \STATE $\mathcal{S} \leftarrow (\mathcal{S} \cup \mathcal{P}) - \mathcal{R}$
    \ENDIF
\ENDIF
\end{algorithmic}
\end{algorithm}

%% file: table_coreset.tex
\begin{table*}[t]
\renewcommand{\thempfootnote}{\fnsymbol{mpfootnote}}
\caption{Statistics of 10 datasets in the proposed benchmark. For KITTI We use the merged set of classes from the universal object detection benchmark~\cite{wang2019towards}.}
\label{tb:coreset}
\begin{minipage}[h]{\linewidth}
\centering
\resizebox{0.85\linewidth}{!}{%
\begin{tabular}{ccccccc}
\toprule
Domain & Dataset & \# classes & \# train images & \# train anno. & \# test images & \# test anno. \cr
\cmidrule(rl){1-1} \cmidrule(rl){2-2} \cmidrule(rl){3-3} \cmidrule(rl){4-4} \cmidrule(rl){5-5} \cmidrule(rl){6-6} \cmidrule(rl){7-7}
Aerial & VisDrone & 10 & 7019 & 381965 & 1610 & 75103 \cr
Agriculture & DeepFruits & 7 & 457 & 2553 & 114 & 590 \cr
Animal & iWildCam & 1 & 21065 & 31591 & 5313 & 7901 \cr
Cartoon & Clipart & 20 & 500 & 1640 & 500 & 1527 \cr
Fashion & iMaterialist & 46 & 45623 & 333402 & 1158 & 8782 \cr
Food & Oktoberfest & 15 & 1110 & 2697 & 85 & 236 \cr
Logo & LogoDet-3K & 352 & 18752 & 35264 & 8331 & 15945 \cr
Person & CrowdHuman & 2 & 15000 & 705967 & 4370 & 206231 \cr
Security & SIXray & 5 & 7496 & 15439 & 1310 & 2054 \cr
Traffic & KITTI & 4 & 5481 & 38077 & 7481 & 52458 \cr
\bottomrule
\end{tabular}
}
\end{minipage}
\end{table*}

%% file: table_arch_1shot.tex
\begin{table*}[t]
\def\smidrulea{\cmidrule(rl){1-2}\cmidrule(rl){3-3}\cmidrule(rl){4-4}\cmidrule(rl){5-5}\cmidrule(rl){6-6}\cmidrule(rl){7-7}\cmidrule(rl){8-8}\cmidrule(rl){9-9}\cmidrule(rl){10-10}\cmidrule(rl){11-11}\cmidrule(rl){12-12}}
\def\smidruleb{\cmidrule(rl){1-1}\cmidrule(rl){2-2}\cmidrule(rl){3-3}\cmidrule(rl){4-4}\cmidrule(rl){5-5}\cmidrule(rl){6-6}\cmidrule(rl){7-7}\cmidrule(rl){8-8}\cmidrule(rl){9-9}\cmidrule(rl){10-10}\cmidrule(rl){11-11}\cmidrule(rl){12-12}\cmidrule(rl){13-13}\cmidrule(rl){14-14}}
\def\smidrulec{\cmidrule(rl){1-1}\cmidrule(rl){2-2}\cmidrule(rl){3-3}\cmidrule(rl){4-4}\cmidrule(rl){5-5}\cmidrule(rl){6-6}\cmidrule(rl){7-7}\cmidrule(rl){8-8}\cmidrule(rl){9-9}\cmidrule(rl){10-10}\cmidrule(rl){11-11}}
\def\smidruled{\cmidrule(rl){1-1}\cmidrule(rl){2-2}\cmidrule(rl){3-3}\cmidrule(rl){4-4}\cmidrule(rl){5-5}\cmidrule(rl){6-6}\cmidrule(rl){7-7}\cmidrule(rl){8-8}\cmidrule(rl){9-9}\cmidrule(rl){10-10}\cmidrule(rl){11-11}\cmidrule(rl){12-12}\cmidrule(rl){13-13}}
\centering
\caption{Per-dataset 1-shot performance of the effects of tuning different parameters, different architectures and pre-training datasets.}
\label{tab:1_shot}
\begin{subtable}[h]{\textwidth}
\resizebox{\linewidth}{!}{
\begin{tabular}{ccccccccccccc}
\toprule
1-shot & Aerial & Agriculture & Animal & Cartoon & Fashion & Food & Logo & Person & Security & Traffic & \multirowcell{2}{Mean} & \multirowcell{2}{Rank} \cr \smidrulec
Unfrozen & VisDrone & DeepFruits & iWildCam & Clipart & iMaterialist & Oktoberfest & LogoDet-3K & CrowdHuman & SIXray & KITTI \cr \smidruled
Last FC layers (TFA~\cite{wang2020frustratingly})  &7.5 \scriptsize{$\pm$ 0.5} & 28.7 \scriptsize{$\pm$ 5.0} & 55.5 \scriptsize{$\pm$ 17.3} & 29.2 \scriptsize{$\pm$ 2.5} & 6.7 \scriptsize{$\pm$ 1.2} & 21.9 \scriptsize{$\pm$ 3.3} & 12.3 \scriptsize{$\pm$ 3.9} & 26.3 \scriptsize{$\pm$ 2.1} & 2.6 \scriptsize{$\pm$ 1.8} & 43.6 \scriptsize{$\pm$ 5.7} & 23.4 \scriptsize{$\pm$ 4.6} & 2.7 \scriptsize{$\pm$ 0.3} \cr
\rowcolor{Gray} Detection Head (FSCE-base~\cite{sun2021fsce}) & 7.8 \scriptsize{$\pm$ 0.8} & 34.6 \scriptsize{$\pm$ 6.3} & 62.3 \scriptsize{$\pm$ 8.2} & 30.6 \scriptsize{$\pm$ 2.6} & 14.1 \scriptsize{$\pm$ 1.2} & 41.0 \scriptsize{$\pm$ 4.0} & 24.1 \scriptsize{$\pm$ 5.5} & 44.4 \scriptsize{$\pm$ 4.8} & 4.5 \scriptsize{$\pm$ 2.4} & 41.3 \scriptsize{$\pm$ 5.5} & 30.5 \scriptsize{$\pm$ 2.3} & 1.9 \scriptsize{$\pm$ 0.1} \cr
Whole Network (Ours-FT) & 8.4 \scriptsize{$\pm$ 1.0} & 36.2 \scriptsize{$\pm$ 7.7} & 56.1 \scriptsize{$\pm$ 5.1} & 37.2 \scriptsize{$\pm$ 3.9} & 14.2 \scriptsize{$\pm$ 1.5} & 47.7 \scriptsize{$\pm$ 6.7} & 25.0 \scriptsize{$\pm$ 4.7} & 45.0 \scriptsize{$\pm$ 4.2} & 6.6 \scriptsize{$\pm$ 4.3} & 38.8 \scriptsize{$\pm$ 3.6} & \textbf{31.5} \scriptsize{$\pm$ 2.0} & \textbf{1.5} \scriptsize{$\pm$ 0.3} \cr
\bottomrule
\end{tabular}
}
\caption{Fine-tuning different number of parameters with Faster R-CNN pre-trained on COCO.
}
\label{tab:tune_1shot}
\end{subtable}

\begin{subtable}[h]{\textwidth}
\resizebox{\linewidth}{!}{
\begin{tabular}{c@{\hskip 6pt}c@{\hskip 6pt}cccccccccccc}
\toprule
\multicolumn{2}{c}{1-shot} & Aerial & Agriculture & Animal & Cartoon & Fashion & Food & Logo & Person & Security & Traffic & \multirowcell{2}{Mean} & \multirowcell{2}{Rank} \cr \smidrulea
Architecture & Pre-training & VisDrone & DeepFruits & iWildCam & Clipart & iMaterialist & Oktoberfest & LogoDet-3K & CrowdHuman & SIXray & KITTI \cr \smidruleb
Faster R-CNN & \cellcolor{White} & 8.4 \scriptsize{$\pm$ 1.0} & 36.2 \scriptsize{$\pm$ 7.7} & 56.1 \scriptsize{$\pm$ 5.1} & 37.2 \scriptsize{$\pm$ 3.9} & 14.2 \scriptsize{$\pm$ 1.5} & 47.7 \scriptsize{$\pm$ 6.7} & 25.0 \scriptsize{$\pm$ 4.7} & 45.0 \scriptsize{$\pm$ 4.2} & 6.6 \scriptsize{$\pm$ 4.3} & 38.8 \scriptsize{$\pm$ 3.6} & 31.5 \scriptsize{$\pm$ 2.0} & 3.6 \scriptsize{$\pm$ 0.2} \cr
\rowcolor{Gray} Cascade R-CNN & \cellcolor{White} & 7.2 \scriptsize{$\pm$ 0.8} & 35.2 \scriptsize{$\pm$ 6.4} & 56.3 \scriptsize{$\pm$ 8.0} & 39.0 \scriptsize{$\pm$ 3.6} & 13.0 \scriptsize{$\pm$ 1.2} & 47.1 \scriptsize{$\pm$ 6.4} & 27.6 \scriptsize{$\pm$ 4.3} & 44.5 \scriptsize{$\pm$ 4.2} & 6.5 \scriptsize{$\pm$ 4.5} & 38.3 \scriptsize{$\pm$ 5.6} & 31.5 \scriptsize{$\pm$ 2.1} & 3.9 \scriptsize{$\pm$ 0.4} \cr
CenterNet2 & \cellcolor{White} & 7.9 \scriptsize{$\pm$ 0.8} & 35.6 \scriptsize{$\pm$ 5.2} & 38.2 \scriptsize{$\pm$ 20.6} & 33.0 \scriptsize{$\pm$ 7.7} & 14.7 \scriptsize{$\pm$ 2.2} & 45.2 \scriptsize{$\pm$ 6.7} & 27.6 \scriptsize{$\pm$ 4.3} & 43.8 \scriptsize{$\pm$ 4.8} & 7.0 \scriptsize{$\pm$ 4.1} & 38.4 \scriptsize{$\pm$ 3.6} & 29.1 \scriptsize{$\pm$ 5.2} & 4.0 \scriptsize{$\pm$ 0.3} \cr
\rowcolor{Gray} RetinaNet & \cellcolor{White} &5.5 \scriptsize{$\pm$ 0.6} & 27.6 \scriptsize{$\pm$ 7.3} & 50.9 \scriptsize{$\pm$ 14.6} & 10.2 \scriptsize{$\pm$ 1.7} & 8.7 \scriptsize{$\pm$ 0.7} & 42.6 \scriptsize{$\pm$ 5.8} & 24.3 \scriptsize{$\pm$ 4.1} & 41.8 \scriptsize{$\pm$ 2.8} & 6.3 \scriptsize{$\pm$ 3.9} & 36.3 \scriptsize{$\pm$ 2.3} & 25.4 \scriptsize{$\pm$ 4.0} & 5.4 \scriptsize{$\pm$ 0.6} \cr
Deformable-DETR& \cellcolor{White} & 8.7 \scriptsize{$\pm$ 1.1} & 44.0 \scriptsize{$\pm$ 6.4} & 53.2 \scriptsize{$\pm$ 12.1} & 23.4 \scriptsize{$\pm$ 4.4} & 15.3 \scriptsize{$\pm$ 1.3} & 47.8 \scriptsize{$\pm$ 6.3} & 28.0 \scriptsize{$\pm$ 4.9} & 47.8 \scriptsize{$\pm$ 4.4} & 10.1 \scriptsize{$\pm$ 4.6} & 41.9 \scriptsize{$\pm$ 5.1} & 32.0 \scriptsize{$\pm$ 2.9} & 2.5 \scriptsize{$\pm$ 0.5} \cr
\rowcolor{Gray}  Cascade R-CNN-P67 & \cellcolor{White} \multirowcell{-6}{COCO} & 9.6 \scriptsize{$\pm$ 1.0} & 41.0 \scriptsize{$\pm$ 4.9} & 65.5 \scriptsize{$\pm$ 7.2} & 44.0 \scriptsize{$\pm$ 2.5} & 16.2 \scriptsize{$\pm$ 1.6} & 51.3 \scriptsize{$\pm$ 5.9} & 28.3 \scriptsize{$\pm$ 4.9} & 47.0 \scriptsize{$\pm$ 3.9} & 8.8 \scriptsize{$\pm$ 4.5} & 42.1 \scriptsize{$\pm$ 4.5} & \textbf{35.4} \scriptsize{$\pm$ 1.8} & \textbf{1.6} \scriptsize{$\pm$ 0.3} \cr
\cmidrule(lr){1-1}\cmidrule(lr){2-2}\cmidrule(lr){3-14}
Faster R-CNN & \cellcolor{Gray} &  8.3 \scriptsize{$\pm$ 0.7} & 45.2 \scriptsize{$\pm$ 4.4} & 58.7 \scriptsize{$\pm$ 6.4} & 24.2 \scriptsize{$\pm$ 3.9} & 20.7 \scriptsize{$\pm$ 1.3} & 49.2 \scriptsize{$\pm$ 5.2} & 25.6 \scriptsize{$\pm$ 5.2} & 41.6 \scriptsize{$\pm$ 3.2} & 8.6 \scriptsize{$\pm$ 3.8} & 34.9 \scriptsize{$\pm$ 3.5} & 31.7 \scriptsize{$\pm$ 1.7} & 2.1 \scriptsize{$\pm$ 0.2} \cr
\rowcolor{Gray} CenterNet2 & \cellcolor{Gray} &  7.6 \scriptsize{$\pm$ 0.7} & 41.5 \scriptsize{$\pm$ 6.0} & 36.0 \scriptsize{$\pm$ 12.7} & 20.0 \scriptsize{$\pm$ 2.4} & 18.7 \scriptsize{$\pm$ 1.4} & 50.4 \scriptsize{$\pm$ 7.1} & 27.8 \scriptsize{$\pm$ 5.0} & 38.7 \scriptsize{$\pm$ 2.6} & 8.6 \scriptsize{$\pm$ 4.2} & 32.3 \scriptsize{$\pm$ 3.6} & 28.1 \scriptsize{$\pm$ 3.3} & 2.7 \scriptsize{$\pm$ 0.3} \cr
Cascade R-CNN-P67 & \cellcolor{Gray} \multirow{-3}{*}{LVIS} & 9.2 \scriptsize{$\pm$ 0.8} & 46.4 \scriptsize{$\pm$ 6.1} & 61.2 \scriptsize{$\pm$ 6.0} & 29.9 \scriptsize{$\pm$ 2.9} & 23.1 \scriptsize{$\pm$ 1.4} & 52.4 \scriptsize{$\pm$ 6.9} & 31.0 \scriptsize{$\pm$ 5.3} & 43.4 \scriptsize{$\pm$ 2.7} & 9.2 \scriptsize{$\pm$ 4.7} & 38.5 \scriptsize{$\pm$ 4.8} & \textbf{34.4} \scriptsize{$\pm$ 2.0} & \textbf{1.2} \scriptsize{$\pm$ 0.2} \cr
\bottomrule
\end{tabular}
}
\caption{Performance of different architectures pre-trained on COCO and LVIS.}
\label{tab:arch_1shot}
\end{subtable}

\begin{subtable}[h]{\textwidth}
\resizebox{\linewidth}{!}{
\begin{tabular}{c@{\hskip 6pt}ccccccccccccc}
\toprule
\multicolumn{2}{c}{1-shot} & Aerial & Agriculture & Animal & Cartoon & Fashion & Food & Logo & Person & Security & Traffic & \multirowcell{2}{Mean} & \multirowcell{2}{Rank} \cr \smidrulea
Architecture & Pre-training & VisDrone & DeepFruits & iWildCam & Clipart & iMaterialist & Oktoberfest & LogoDet-3K & CrowdHuman & SIXray & KITTI \cr \smidruleb
& ImageNet &4.8 \scriptsize{$\pm$ 0.5} & 23.5 \scriptsize{$\pm$ 4.5} & 1.0 \scriptsize{$\pm$ 0.8} & 2.8 \scriptsize{$\pm$ 1.0} & 9.3 \scriptsize{$\pm$ 1.1} & 43.3 \scriptsize{$\pm$ 5.2} & 23.1 \scriptsize{$\pm$ 3.8} & 14.8 \scriptsize{$\pm$ 2.5} & 2.1 \scriptsize{$\pm$ 1.3} & 10.7 \scriptsize{$\pm$ 2.7} & 13.5 \scriptsize{$\pm$ 1.6} & 6.0 \scriptsize{$\pm$ 0.1} \cr
\rowcolor{Gray} \cellcolor{White} & COCO & 9.6 \scriptsize{$\pm$ 1.0} & 41.0 \scriptsize{$\pm$ 4.9} & 65.5 \scriptsize{$\pm$ 7.2} & 44.0 \scriptsize{$\pm$ 2.5} & 16.2 \scriptsize{$\pm$ 1.6} & 51.3 \scriptsize{$\pm$ 5.9} & 28.3 \scriptsize{$\pm$ 4.9} & 47.0 \scriptsize{$\pm$ 3.9} & 8.8 \scriptsize{$\pm$ 4.5} & 42.1 \scriptsize{$\pm$ 4.5} & \textbf{35.4} \scriptsize{$\pm$ 1.8} & 2.9 \scriptsize{$\pm$ 0.4} \cr
\cellcolor{White} &FSODD & 5.8 \scriptsize{$\pm$ 0.5} & 48.0 \scriptsize{$\pm$ 3.9} & 44.1 \scriptsize{$\pm$ 11.1} & 12.5 \scriptsize{$\pm$ 2.7} & 14.8 \scriptsize{$\pm$ 1.2} & 50.1 \scriptsize{$\pm$ 6.5} & 29.3 \scriptsize{$\pm$ 4.2} & 28.6 \scriptsize{$\pm$ 1.8} & 8.0 \scriptsize{$\pm$ 3.6} & 25.5 \scriptsize{$\pm$ 3.0} & 26.7 \scriptsize{$\pm$ 2.9} & 4.2 \scriptsize{$\pm$ 0.5} \cr
\rowcolor{Gray} \cellcolor{White} & LVIS & 9.2 \scriptsize{$\pm$ 0.8} & 46.4 \scriptsize{$\pm$ 6.1} & 61.2 \scriptsize{$\pm$ 6.0} & 29.9 \scriptsize{$\pm$ 2.9} & 23.1 \scriptsize{$\pm$ 1.4} & 52.4 \scriptsize{$\pm$ 6.9} & 31.0 \scriptsize{$\pm$ 5.3} & 43.4 \scriptsize{$\pm$ 2.7} & 9.2 \scriptsize{$\pm$ 4.7} & 38.5 \scriptsize{$\pm$ 4.8} & 34.4 \scriptsize{$\pm$ 2.0} & 3.0 \scriptsize{$\pm$ 0.2} \cr
\cellcolor{White} & Unified & 9.7 \scriptsize{$\pm$ 1.0} & 47.2 \scriptsize{$\pm$ 6.5} & 45.8 \scriptsize{$\pm$ 8.5} & 31.2 \scriptsize{$\pm$ 5.1} & 18.4 \scriptsize{$\pm$ 1.3} & 52.8 \scriptsize{$\pm$ 6.1} & 31.2 \scriptsize{$\pm$ 5.1} & 46.4 \scriptsize{$\pm$ 3.0} & 10.4 \scriptsize{$\pm$ 4.8} & 39.5 \scriptsize{$\pm$ 3.6} & 33.3 \scriptsize{$\pm$ 2.2} & 2.8 \scriptsize{$\pm$ 0.3} \cr
\rowcolor{Gray} \cellcolor{White} \multirow{-6}{*}{Cascade R-CNN-P67~} & LVIS+ & 
11.7 \scriptsize{$\pm$ 1.0} & 57.4 \scriptsize{$\pm$ 6.4} & 30.9 \scriptsize{$\pm$ 15.9} & 37.3 \scriptsize{$\pm$ 1.9} & 25.5 \scriptsize{$\pm$ 0.9} & 50.0 \scriptsize{$\pm$ 8.0} & 36.0 \scriptsize{$\pm$ 3.8} & 45.1 \scriptsize{$\pm$ 3.1} & 13.3 \scriptsize{$\pm$ 5.2} & 39.5 \scriptsize{$\pm$ 4.9} & 34.7 \scriptsize{$\pm$ 4.2} & \textbf{2.1} \scriptsize{$\pm$ 0.5} \cr
\cmidrule(lr){1-1}\cmidrule(lr){2-2}\cmidrule(lr){3-14}
\cellcolor{Gray} & COCO & 7.9 \scriptsize{$\pm$ 0.8} & 35.6 \scriptsize{$\pm$ 5.2} & 38.2 \scriptsize{$\pm$ 20.6} & 33.0 \scriptsize{$\pm$ 7.7} & 14.7 \scriptsize{$\pm$ 2.2} & 45.2 \scriptsize{$\pm$ 6.7} & 27.6 \scriptsize{$\pm$ 4.3} & 43.8 \scriptsize{$\pm$ 4.8} & 7.0 \scriptsize{$\pm$ 4.1} & 38.4 \scriptsize{$\pm$ 3.6} & 29.1 \scriptsize{$\pm$ 5.2} & 3.1 \scriptsize{$\pm$ 0.4} \cr
\rowcolor{Gray} & LVIS & 7.6 \scriptsize{$\pm$ 0.7} & 41.5 \scriptsize{$\pm$ 6.0} & 36.0 \scriptsize{$\pm$ 12.7} & 20.0 \scriptsize{$\pm$ 2.4} & 18.7 \scriptsize{$\pm$ 1.4} & 50.4 \scriptsize{$\pm$ 7.1} & 27.8 \scriptsize{$\pm$ 5.0} & 38.7 \scriptsize{$\pm$ 2.6} & 8.6 \scriptsize{$\pm$ 4.2} & 32.3 \scriptsize{$\pm$ 3.6} & 28.1 \scriptsize{$\pm$ 3.3} & 3.3 \scriptsize{$\pm$ 0.3} \cr
\cellcolor{Gray} & LVIS+ & 10.6 \scriptsize{$\pm$ 1.1} & 55.7 \scriptsize{$\pm$ 5.4} & 38.4 \scriptsize{$\pm$ 12.8} & 35.5 \scriptsize{$\pm$ 2.9} & 25.1 \scriptsize{$\pm$ 1.1} & 48.4 \scriptsize{$\pm$ 6.7} & 36.7 \scriptsize{$\pm$ 4.2} & 46.4 \scriptsize{$\pm$ 3.3} & 13.9 \scriptsize{$\pm$ 5.7} & 37.9 \scriptsize{$\pm$ 5.3} & 34.9 \scriptsize{$\pm$ 3.2} & 1.9 \scriptsize{$\pm$ 0.3} \cr
\rowcolor{Gray} \multirow{-4}{*}{CenterNet2} & LVIS++ & 10.7 \scriptsize{$\pm$ 1.0} & 59.4 \scriptsize{$\pm$ 5.5} & 41.7 \scriptsize{$\pm$ 13.4} & 38.2 \scriptsize{$\pm$ 2.3} & 26.7 \scriptsize{$\pm$ 1.0} & 46.2 \scriptsize{$\pm$ 6.1} & 35.6 \scriptsize{$\pm$ 4.2} & 47.0 \scriptsize{$\pm$ 3.5} & 15.7 \scriptsize{$\pm$ 5.8} & 37.1 \scriptsize{$\pm$ 4.7} & \textbf{35.8} \scriptsize{$\pm$ 3.4} & \textbf{1.7} \scriptsize{$\pm$ 0.3} \cr
\bottomrule
\end{tabular}
}
\caption{Performance of Cascade R-CNN-P67 and CenterNet2 pre-trained on different datasets.}
\label{tab:pre_training_1shot}
\end{subtable}

\end{table*}

%% file: table_arch_3shot.tex
\begin{table*}[t]
\def\smidrulea{\cmidrule(rl){1-2}\cmidrule(rl){3-3}\cmidrule(rl){4-4}\cmidrule(rl){5-5}\cmidrule(rl){6-6}\cmidrule(rl){7-7}\cmidrule(rl){8-8}\cmidrule(rl){9-9}\cmidrule(rl){10-10}\cmidrule(rl){11-11}\cmidrule(rl){12-12}}
\def\smidruleb{\cmidrule(rl){1-1}\cmidrule(rl){2-2}\cmidrule(rl){3-3}\cmidrule(rl){4-4}\cmidrule(rl){5-5}\cmidrule(rl){6-6}\cmidrule(rl){7-7}\cmidrule(rl){8-8}\cmidrule(rl){9-9}\cmidrule(rl){10-10}\cmidrule(rl){11-11}\cmidrule(rl){12-12}\cmidrule(rl){13-13}\cmidrule(rl){14-14}}
\def\smidrulec{\cmidrule(rl){1-1}\cmidrule(rl){2-2}\cmidrule(rl){3-3}\cmidrule(rl){4-4}\cmidrule(rl){5-5}\cmidrule(rl){6-6}\cmidrule(rl){7-7}\cmidrule(rl){8-8}\cmidrule(rl){9-9}\cmidrule(rl){10-10}\cmidrule(rl){11-11}}
\def\smidruled{\cmidrule(rl){1-1}\cmidrule(rl){2-2}\cmidrule(rl){3-3}\cmidrule(rl){4-4}\cmidrule(rl){5-5}\cmidrule(rl){6-6}\cmidrule(rl){7-7}\cmidrule(rl){8-8}\cmidrule(rl){9-9}\cmidrule(rl){10-10}\cmidrule(rl){11-11}\cmidrule(rl){12-12}\cmidrule(rl){13-13}}
\centering
\caption{Per-dataset 3-shot performance of the effects of tuning different parameters, different architectures and pre-training datasets.}
\label{tab:3_shot}
\begin{subtable}[h]{\textwidth}
\resizebox{\linewidth}{!}{
\begin{tabular}{ccccccccccccc}
\toprule
3-shot & Aerial & Agriculture & Animal & Cartoon & Fashion & Food & Logo & Person & Security & Traffic & \multirowcell{2}{Mean} & \multirowcell{2}{Rank} \cr \smidrulec
Unfrozen & VisDrone & DeepFruits & iWildCam & Clipart & iMaterialist & Oktoberfest & LogoDet-3K & CrowdHuman & SIXray & KITTI \cr \smidruled
Last FC layers (TFA~\cite{wang2020frustratingly})  &9.4 \scriptsize{$\pm$ 0.4} & 41.8 \scriptsize{$\pm$ 3.0} & 70.3 \scriptsize{$\pm$ 4.5} & 35.9 \scriptsize{$\pm$ 2.3} & 7.7 \scriptsize{$\pm$ 1.3} & 32.0 \scriptsize{$\pm$ 6.9} & 13.3 \scriptsize{$\pm$ 3.3} & 29.6 \scriptsize{$\pm$ 1.0} & 5.4 \scriptsize{$\pm$ 1.9} & 46.6 \scriptsize{$\pm$ 4.2} & 29.2 \scriptsize{$\pm$ 1.8} & 2.7 \scriptsize{$\pm$ 0.2} \cr
\rowcolor{Gray} Detection Head (FSCE-base~\cite{sun2021fsce}) & 11.4 \scriptsize{$\pm$ 0.7} & 51.6 \scriptsize{$\pm$ 4.8} & 70.0 \scriptsize{$\pm$ 1.9} & 38.7 \scriptsize{$\pm$ 1.7} & 18.4 \scriptsize{$\pm$ 1.0} & 61.4 \scriptsize{$\pm$ 5.4} & 38.4 \scriptsize{$\pm$ 4.8} & 49.0 \scriptsize{$\pm$ 3.2} & 10.7 \scriptsize{$\pm$ 3.5} & 44.6 \scriptsize{$\pm$ 4.4} & 39.4 \scriptsize{$\pm$ 1.6} & 1.9 \scriptsize{$\pm$ 0.2} \cr
Whole Network (Ours-FT) & 12.0 \scriptsize{$\pm$ 0.8} & 52.6 \scriptsize{$\pm$ 4.6} & 62.9 \scriptsize{$\pm$ 5.2} & 45.5 \scriptsize{$\pm$ 3.1} & 19.3 \scriptsize{$\pm$ 1.0} & 70.1 \scriptsize{$\pm$ 5.9} & 41.7 \scriptsize{$\pm$ 4.5} & 48.8 \scriptsize{$\pm$ 2.7} & 15.5 \scriptsize{$\pm$ 6.2} & 43.2 \scriptsize{$\pm$ 3.6} & \textbf{41.1} \scriptsize{$\pm$ 1.8} & \textbf{1.5} \scriptsize{$\pm$ 0.2} \cr
\bottomrule
\end{tabular}
}
\caption{Fine-tuning different number of parameters with Faster R-CNN pre-trained on COCO.
}
\label{tab:tune_3shot}
\end{subtable}

\begin{subtable}[h]{\textwidth}
\resizebox{\linewidth}{!}{
\begin{tabular}{c@{\hskip 6pt}c@{\hskip 6pt}cccccccccccc}
\toprule
\multicolumn{2}{c}{3-shot} & Aerial & Agriculture & Animal & Cartoon & Fashion & Food & Logo & Person & Security & Traffic & \multirowcell{2}{Mean} & \multirowcell{2}{Rank} \cr \smidrulea
Architecture & Pre-training & VisDrone & DeepFruits & iWildCam & Clipart & iMaterialist & Oktoberfest & LogoDet-3K & CrowdHuman & SIXray & KITTI \cr \smidruleb
Faster R-CNN & \cellcolor{White} & 12.0 \scriptsize{$\pm$ 0.8} & 52.6 \scriptsize{$\pm$ 4.6} & 62.9 \scriptsize{$\pm$ 5.2} & 45.5 \scriptsize{$\pm$ 3.1} & 19.3 \scriptsize{$\pm$ 1.0} & 70.1 \scriptsize{$\pm$ 5.9} & 41.7 \scriptsize{$\pm$ 4.5} & 48.8 \scriptsize{$\pm$ 2.7} & 15.5 \scriptsize{$\pm$ 6.2} & 43.2 \scriptsize{$\pm$ 3.6} & 41.1 \scriptsize{$\pm$ 1.8} & 3.5 \scriptsize{$\pm$ 0.3} \cr
\rowcolor{Gray} Cascade R-CNN & \cellcolor{White} & 11.0 \scriptsize{$\pm$ 0.7} & 52.3 \scriptsize{$\pm$ 2.9} & 67.7 \scriptsize{$\pm$ 3.2} & 45.9 \scriptsize{$\pm$ 2.5} & 18.3 \scriptsize{$\pm$ 0.9} & 69.5 \scriptsize{$\pm$ 5.1} & 42.3 \scriptsize{$\pm$ 4.1} & 48.9 \scriptsize{$\pm$ 2.6} & 14.4 \scriptsize{$\pm$ 5.9} & 41.8 \scriptsize{$\pm$ 3.1} & 41.2 \scriptsize{$\pm$ 1.5} & 3.8 \scriptsize{$\pm$ 0.3} \cr
CenterNet2 & \cellcolor{White} & 11.6 \scriptsize{$\pm$ 0.7} & 50.9 \scriptsize{$\pm$ 6.3} & 60.6 \scriptsize{$\pm$ 4.9} & 44.0 \scriptsize{$\pm$ 6.8} & 19.7 \scriptsize{$\pm$ 2.4} & 68.2 \scriptsize{$\pm$ 5.9} & 42.7 \scriptsize{$\pm$ 2.4} & 48.6 \scriptsize{$\pm$ 3.9} & 15.5 \scriptsize{$\pm$ 5.7} & 40.5 \scriptsize{$\pm$ 4.4} & 40.2 \scriptsize{$\pm$ 1.9} & 3.8 \scriptsize{$\pm$ 0.4} \cr
\rowcolor{Gray} RetinaNet & \cellcolor{White} &8.2 \scriptsize{$\pm$ 0.5} & 45.7 \scriptsize{$\pm$ 3.2} & 59.0 \scriptsize{$\pm$ 7.2} & 19.2 \scriptsize{$\pm$ 1.8} & 14.5 \scriptsize{$\pm$ 0.6} & 66.5 \scriptsize{$\pm$ 6.4} & 39.1 \scriptsize{$\pm$ 4.7} & 45.5 \scriptsize{$\pm$ 2.0} & 12.0 \scriptsize{$\pm$ 4.9} & 37.8 \scriptsize{$\pm$ 2.9} & 34.8 \scriptsize{$\pm$ 2.2} & 5.6 \scriptsize{$\pm$ 0.6} \cr
Deformable-DETR& \cellcolor{White} & 12.7 \scriptsize{$\pm$ 0.7} & 61.1 \scriptsize{$\pm$ 4.3} & 64.8 \scriptsize{$\pm$ 3.2} & 35.9 \scriptsize{$\pm$ 2.7} & 19.9 \scriptsize{$\pm$ 1.2} & 67.9 \scriptsize{$\pm$ 4.6} & 42.5 \scriptsize{$\pm$ 4.6} & 53.1 \scriptsize{$\pm$ 3.1} & 21.3 \scriptsize{$\pm$ 6.7} & 43.4 \scriptsize{$\pm$ 3.5} & 42.3 \scriptsize{$\pm$ 1.6} & 2.7 \scriptsize{$\pm$ 0.5} \cr
\rowcolor{Gray} Cascade R-CNN-P67 & \cellcolor{White} \multirowcell{-6}{COCO} & 13.7 \scriptsize{$\pm$ 0.9} & 55.3 \scriptsize{$\pm$ 3.0} & 72.8 \scriptsize{$\pm$ 2.5} & 52.1 \scriptsize{$\pm$ 2.4} & 21.9 \scriptsize{$\pm$ 1.0} & 71.2 \scriptsize{$\pm$ 5.4} & 46.6 \scriptsize{$\pm$ 4.4} & 51.2 \scriptsize{$\pm$ 2.4} & 17.9 \scriptsize{$\pm$ 5.6} & 44.4 \scriptsize{$\pm$ 2.8} & \textbf{44.7} \scriptsize{$\pm$ 1.6} & \textbf{1.7} \scriptsize{$\pm$ 0.5} \cr
\cmidrule(lr){1-1}\cmidrule(lr){2-2}\cmidrule(lr){3-14}
Faster R-CNN & \cellcolor{Gray} &  11.9 \scriptsize{$\pm$ 0.6} & 59.2 \scriptsize{$\pm$ 5.2} & 69.4 \scriptsize{$\pm$ 3.1} & 33.8 \scriptsize{$\pm$ 3.5} & 25.8 \scriptsize{$\pm$ 0.9} & 71.1 \scriptsize{$\pm$ 4.4} & 41.5 \scriptsize{$\pm$ 4.2} & 45.9 \scriptsize{$\pm$ 2.5} & 18.3 \scriptsize{$\pm$ 5.1} & 38.6 \scriptsize{$\pm$ 3.6} & 41.6 \scriptsize{$\pm$ 1.5} & 2.1 \scriptsize{$\pm$ 0.3} \cr
\rowcolor{Gray} CenterNet2 & \cellcolor{Gray} &  11.2 \scriptsize{$\pm$ 0.6} & 56.9 \scriptsize{$\pm$ 3.9} & 59.0 \scriptsize{$\pm$ 5.5} & 27.9 \scriptsize{$\pm$ 3.9} & 23.1 \scriptsize{$\pm$ 0.6} & 70.7 \scriptsize{$\pm$ 5.0} & 45.3 \scriptsize{$\pm$ 4.0} & 43.8 \scriptsize{$\pm$ 2.3} & 16.5 \scriptsize{$\pm$ 5.3} & 35.8 \scriptsize{$\pm$ 4.2} & 39.0 \scriptsize{$\pm$ 1.7} & 2.7 \scriptsize{$\pm$ 0.3} \cr
Cascade R-CNN-P67 & \cellcolor{Gray} \multirow{-3}{*}{LVIS} & 13.1 \scriptsize{$\pm$ 0.7} & 59.9 \scriptsize{$\pm$ 5.3} & 71.1 \scriptsize{$\pm$ 3.1} & 39.6 \scriptsize{$\pm$ 3.7} & 28.1 \scriptsize{$\pm$ 1.0} & 71.9 \scriptsize{$\pm$ 5.5} & 47.7 \scriptsize{$\pm$ 2.8} & 47.3 \scriptsize{$\pm$ 2.4} & 19.0 \scriptsize{$\pm$ 5.4} & 42.8 \scriptsize{$\pm$ 4.1} & \textbf{44.0} \scriptsize{$\pm$ 1.6} & \textbf{1.2} \scriptsize{$\pm$ 0.3} \cr
\bottomrule
\end{tabular}
}
\caption{Performance of different architectures pre-trained on COCO and LVIS.}
\label{tab:arch_3shot}
\end{subtable}

\begin{subtable}[h]{\textwidth}
\resizebox{\linewidth}{!}{
\begin{tabular}{c@{\hskip 6pt}ccccccccccccc}
\toprule
\multicolumn{2}{c}{3-shot} & Aerial & Agriculture & Animal & Cartoon & Fashion & Food & Logo & Person & Security & Traffic & \multirowcell{2}{Mean} & \multirowcell{2}{Rank} \cr \smidrulea
Architecture & Pre-training & VisDrone & DeepFruits & iWildCam & Clipart & iMaterialist & Oktoberfest & LogoDet-3K & CrowdHuman & SIXray & KITTI \cr \smidruleb
& ImageNet &7.9 \scriptsize{$\pm$ 0.5} & 43.1 \scriptsize{$\pm$ 3.3} & 4.1 \scriptsize{$\pm$ 2.2} & 7.3 \scriptsize{$\pm$ 1.8} & 16.7 \scriptsize{$\pm$ 0.7} & 65.8 \scriptsize{$\pm$ 5.5} & 37.6 \scriptsize{$\pm$ 4.7} & 25.6 \scriptsize{$\pm$ 3.3} & 6.5 \scriptsize{$\pm$ 3.4} & 17.9 \scriptsize{$\pm$ 2.0} & 23.2 \scriptsize{$\pm$ 1.5} & 6.0 \scriptsize{$\pm$ 0.1} \cr
\rowcolor{Gray} \cellcolor{White} & COCO & 13.7 \scriptsize{$\pm$ 0.9} & 55.3 \scriptsize{$\pm$ 3.0} & 72.8 \scriptsize{$\pm$ 2.5} & 52.1 \scriptsize{$\pm$ 2.4} & 21.9 \scriptsize{$\pm$ 1.0} & 71.2 \scriptsize{$\pm$ 5.4} & 46.6 \scriptsize{$\pm$ 4.4} & 51.2 \scriptsize{$\pm$ 2.4} & 17.9 \scriptsize{$\pm$ 5.6} & 44.4 \scriptsize{$\pm$ 2.8} & 44.7 \scriptsize{$\pm$ 1.6} & 2.9 \scriptsize{$\pm$ 0.4} \cr
\cellcolor{White} &FSODD & 9.0 \scriptsize{$\pm$ 0.8} & 61.0 \scriptsize{$\pm$ 3.4} & 62.3 \scriptsize{$\pm$ 4.6} & 19.6 \scriptsize{$\pm$ 2.5} & 19.5 \scriptsize{$\pm$ 0.8} & 71.8 \scriptsize{$\pm$ 5.0} & 46.9 \scriptsize{$\pm$ 4.4} & 35.0 \scriptsize{$\pm$ 2.5} & 16.1 \scriptsize{$\pm$ 5.0} & 28.3 \scriptsize{$\pm$ 3.2} & 36.9 \scriptsize{$\pm$ 1.5} & 4.3 \scriptsize{$\pm$ 0.4} \cr
\rowcolor{Gray} \cellcolor{White} & LVIS & 13.1 \scriptsize{$\pm$ 0.7} & 59.9 \scriptsize{$\pm$ 5.3} & 71.1 \scriptsize{$\pm$ 3.1} & 39.6 \scriptsize{$\pm$ 3.7} & 28.1 \scriptsize{$\pm$ 1.0} & 71.9 \scriptsize{$\pm$ 5.5} & 47.7 \scriptsize{$\pm$ 2.8} & 47.3 \scriptsize{$\pm$ 2.4} & 19.0 \scriptsize{$\pm$ 5.4} & 42.8 \scriptsize{$\pm$ 4.1} & 44.0 \scriptsize{$\pm$ 1.6} & 3.2 \scriptsize{$\pm$ 0.5} \cr
\cellcolor{White} & Unified & 14.0 \scriptsize{$\pm$ 0.9} & 62.5 \scriptsize{$\pm$ 3.2} & 63.7 \scriptsize{$\pm$ 4.2} & 41.9 \scriptsize{$\pm$ 3.3} & 24.2 \scriptsize{$\pm$ 0.8} & 73.7 \scriptsize{$\pm$ 5.1} & 50.1 \scriptsize{$\pm$ 4.5} & 50.3 \scriptsize{$\pm$ 2.2} & 19.7 \scriptsize{$\pm$ 4.5} & 43.1 \scriptsize{$\pm$ 3.0} & 44.3 \scriptsize{$\pm$ 1.4} & 2.7 \scriptsize{$\pm$ 0.4} \cr
\rowcolor{Gray} \cellcolor{White} \multirow{-6}{*}{Cascade R-CNN-P67~} & LVIS+ & 16.3 \scriptsize{$\pm$ 0.9} & 70.7 \scriptsize{$\pm$ 3.5} & 55.0 \scriptsize{$\pm$ 6.3} & 46.8 \scriptsize{$\pm$ 2.3} & 29.8 \scriptsize{$\pm$ 0.7} & 72.1 \scriptsize{$\pm$ 3.5} & 52.2 \scriptsize{$\pm$ 3.6} & 50.1 \scriptsize{$\pm$ 2.6} & 26.8 \scriptsize{$\pm$ 4.9} & 46.0 \scriptsize{$\pm$ 3.4} & \textbf{46.6} \scriptsize{$\pm$ 1.6} & \textbf{1.9} \scriptsize{$\pm$ 0.5} \cr
\cmidrule(lr){1-1}\cmidrule(lr){2-2}\cmidrule(lr){3-14}
\cellcolor{Gray} & COCO & 7.9 \scriptsize{$\pm$ 0.8} & 35.6 \scriptsize{$\pm$ 5.2} & 38.2 \scriptsize{$\pm$ 20.6} & 33.0 \scriptsize{$\pm$ 7.7} & 14.7 \scriptsize{$\pm$ 2.2} & 45.2 \scriptsize{$\pm$ 6.7} & 27.6 \scriptsize{$\pm$ 4.3} & 43.8 \scriptsize{$\pm$ 4.8} & 7.0 \scriptsize{$\pm$ 4.1} & 38.4 \scriptsize{$\pm$ 3.6} & 29.1 \scriptsize{$\pm$ 5.2} & 3.1 \scriptsize{$\pm$ 0.4} \cr
\rowcolor{Gray} & LVIS & 7.6 \scriptsize{$\pm$ 0.7} & 41.5 \scriptsize{$\pm$ 6.0} & 36.0 \scriptsize{$\pm$ 12.7} & 20.0 \scriptsize{$\pm$ 2.4} & 18.7 \scriptsize{$\pm$ 1.4} & 50.4 \scriptsize{$\pm$ 7.1} & 27.8 \scriptsize{$\pm$ 5.0} & 38.7 \scriptsize{$\pm$ 2.6} & 8.6 \scriptsize{$\pm$ 4.2} & 32.3 \scriptsize{$\pm$ 3.6} & 28.1 \scriptsize{$\pm$ 3.3} & 3.3 \scriptsize{$\pm$ 0.3} \cr
\cellcolor{Gray} & LVIS+ & 10.6 \scriptsize{$\pm$ 1.1} & 55.7 \scriptsize{$\pm$ 5.4} & 38.4 \scriptsize{$\pm$ 12.8} & 35.5 \scriptsize{$\pm$ 2.9} & 25.1 \scriptsize{$\pm$ 1.1} & 48.4 \scriptsize{$\pm$ 6.7} & 36.7 \scriptsize{$\pm$ 4.2} & 46.4 \scriptsize{$\pm$ 3.3} & 13.9 \scriptsize{$\pm$ 5.7} & 37.9 \scriptsize{$\pm$ 5.3} & 34.9 \scriptsize{$\pm$ 3.2} & 1.9 \scriptsize{$\pm$ 0.3} \cr
\rowcolor{Gray} \multirow{-4}{*}{CenterNet2} & LVIS++ & 10.7 \scriptsize{$\pm$ 1.0} & 59.4 \scriptsize{$\pm$ 5.5} & 41.7 \scriptsize{$\pm$ 13.4} & 38.2 \scriptsize{$\pm$ 2.3} & 26.7 \scriptsize{$\pm$ 1.0} & 46.2 \scriptsize{$\pm$ 6.1} & 35.6 \scriptsize{$\pm$ 4.2} & 47.0 \scriptsize{$\pm$ 3.5} & 15.7 \scriptsize{$\pm$ 5.8} & 37.1 \scriptsize{$\pm$ 4.7} & \textbf{35.8} \scriptsize{$\pm$ 3.4} & \textbf{1.7} \scriptsize{$\pm$ 0.3} \cr
\bottomrule
\end{tabular}
}
\caption{Performance of Cascade R-CNN-P67 and CenterNet2 pre-trained on different datasets.}
\label{tab:pre_training_3shot}
\end{subtable}

\end{table*}

%% file: table_arch_10shot.tex
\begin{table*}[t]
\def\smidrulea{\cmidrule(rl){1-2}\cmidrule(rl){3-3}\cmidrule(rl){4-4}\cmidrule(rl){5-5}\cmidrule(rl){6-6}\cmidrule(rl){7-7}\cmidrule(rl){8-8}\cmidrule(rl){9-9}\cmidrule(rl){10-10}\cmidrule(rl){11-11}\cmidrule(rl){12-12}}
\def\smidruleb{\cmidrule(rl){1-1}\cmidrule(rl){2-2}\cmidrule(rl){3-3}\cmidrule(rl){4-4}\cmidrule(rl){5-5}\cmidrule(rl){6-6}\cmidrule(rl){7-7}\cmidrule(rl){8-8}\cmidrule(rl){9-9}\cmidrule(rl){10-10}\cmidrule(rl){11-11}\cmidrule(rl){12-12}\cmidrule(rl){13-13}\cmidrule(rl){14-14}}
\def\smidrulec{\cmidrule(rl){1-1}\cmidrule(rl){2-2}\cmidrule(rl){3-3}\cmidrule(rl){4-4}\cmidrule(rl){5-5}\cmidrule(rl){6-6}\cmidrule(rl){7-7}\cmidrule(rl){8-8}\cmidrule(rl){9-9}\cmidrule(rl){10-10}\cmidrule(rl){11-11}}
\def\smidruled{\cmidrule(rl){1-1}\cmidrule(rl){2-2}\cmidrule(rl){3-3}\cmidrule(rl){4-4}\cmidrule(rl){5-5}\cmidrule(rl){6-6}\cmidrule(rl){7-7}\cmidrule(rl){8-8}\cmidrule(rl){9-9}\cmidrule(rl){10-10}\cmidrule(rl){11-11}\cmidrule(rl){12-12}\cmidrule(rl){13-13}}
\centering
\caption{Per-dataset 10-shot performance of the effects of tuning different parameters, different architectures and pre-training datasets.}
\label{tab:10_shot}
\begin{subtable}[h]{\textwidth}
\resizebox{\linewidth}{!}{
\begin{tabular}{ccccccccccccc}
\toprule
10-shot & Aerial & Agriculture & Animal & Cartoon & Fashion & Food & Logo & Person & Security & Traffic & \multirowcell{2}{Mean} & \multirowcell{2}{Rank} \cr \smidrulec
Unfrozen & VisDrone & DeepFruits & iWildCam & Clipart & iMaterialist & Oktoberfest & LogoDet-3K & CrowdHuman & SIXray & KITTI \cr \smidruled
Last FC layers (TFA~\cite{wang2020frustratingly})  & 10.8 \scriptsize{$\pm$ 0.5} & 55.9 \scriptsize{$\pm$ 1.8} & 73.8 \scriptsize{$\pm$ 2.9} & 44.5 \scriptsize{$\pm$ 1.0} & 8.1 \scriptsize{$\pm$ 1.2} & 48.0 \scriptsize{$\pm$ 2.7} & 15.3 \scriptsize{$\pm$ 4.4} & 31.4 \scriptsize{$\pm$ 0.7} & 11.0 \scriptsize{$\pm$ 1.2} & 53.2 \scriptsize{$\pm$ 2.4} & 35.2 \scriptsize{$\pm$ 1.2} & 2.7 \scriptsize{$\pm$ 0.2} \cr
\rowcolor{Gray} Detection Head (FSCE-base~\cite{sun2021fsce}) & 15.5 \scriptsize{$\pm$ 0.6} & 69.7 \scriptsize{$\pm$ 2.8} & 72.2 \scriptsize{$\pm$ 1.8} & 50.2 \scriptsize{$\pm$ 1.1} & 23.2 \scriptsize{$\pm$ 1.2} & 83.0 \scriptsize{$\pm$ 3.3} & 55.7 \scriptsize{$\pm$ 4.3} & 54.4 \scriptsize{$\pm$ 1.8} & 23.8 \scriptsize{$\pm$ 1.9} & 54.0 \scriptsize{$\pm$ 2.5} & 50.2 \scriptsize{$\pm$ 1.1} & 1.8 \scriptsize{$\pm$ 0.2} \cr
Whole Network (Ours-FT) & 17.5 \scriptsize{$\pm$ 0.7} & 71.4 \scriptsize{$\pm$ 1.9} & 65.3 \scriptsize{$\pm$ 6.9} & 57.4 \scriptsize{$\pm$ 1.1} & 24.8 \scriptsize{$\pm$ 1.0} & 90.6 \scriptsize{$\pm$ 1.9} & 59.2 \scriptsize{$\pm$ 4.4} & 53.3 \scriptsize{$\pm$ 1.7} & 36.1 \scriptsize{$\pm$ 3.0} & 50.6 \scriptsize{$\pm$ 3.1} & \textbf{52.6} \scriptsize{$\pm$ 1.8} & \textbf{1.5} \scriptsize{$\pm$ 0.2} \cr
\bottomrule
\end{tabular}
}
\caption{Fine-tuning different number of parameters with Faster R-CNN pre-trained on COCO.
}
\label{tab:tune_10shot}
\end{subtable}

\begin{subtable}[h]{\textwidth}
\resizebox{\linewidth}{!}{
\begin{tabular}{c@{\hskip 6pt}c@{\hskip 6pt}cccccccccccc}
\toprule
\multicolumn{2}{c}{10-shot} & Aerial & Agriculture & Animal & Cartoon & Fashion & Food & Logo & Person & Security & Traffic & \multirowcell{2}{Mean} & \multirowcell{2}{Rank} \cr \smidrulea
Architecture & Pre-training & VisDrone & DeepFruits & iWildCam & Clipart & iMaterialist & Oktoberfest & LogoDet-3K & CrowdHuman & SIXray & KITTI \cr \smidruleb
Faster R-CNN & \cellcolor{White} & 17.5 \scriptsize{$\pm$ 0.7} & 71.4 \scriptsize{$\pm$ 1.9} & 65.3 \scriptsize{$\pm$ 6.9} & 57.4 \scriptsize{$\pm$ 1.1} & 24.8 \scriptsize{$\pm$ 1.0} & 90.6 \scriptsize{$\pm$ 1.9} & 59.2 \scriptsize{$\pm$ 4.4} & 53.3 \scriptsize{$\pm$ 1.7} & 36.1 \scriptsize{$\pm$ 3.0} & 50.6 \scriptsize{$\pm$ 3.1} & 52.6 \scriptsize{$\pm$ 1.8} & 3.8 \scriptsize{$\pm$ 0.4} \cr
\rowcolor{Gray} Cascade R-CNN & \cellcolor{White} & 16.6 \scriptsize{$\pm$ 0.7} & 71.6 \scriptsize{$\pm$ 2.2} & 69.7 \scriptsize{$\pm$ 3.5} & 58.5 \scriptsize{$\pm$ 1.5} & 23.8 \scriptsize{$\pm$ 0.9} & 90.0 \scriptsize{$\pm$ 2.2} & 58.8 \scriptsize{$\pm$ 4.6} & 53.5 \scriptsize{$\pm$ 1.6} & 33.7 \scriptsize{$\pm$ 2.7} & 50.8 \scriptsize{$\pm$ 2.9} & 52.7 \scriptsize{$\pm$ 1.2} & 3.9 \scriptsize{$\pm$ 0.4} \cr
CenterNet2 & \cellcolor{White} & 16.4 \scriptsize{$\pm$ 0.7} & 71.3 \scriptsize{$\pm$ 2.0} & 65.2 \scriptsize{$\pm$ 5.1} & 57.6 \scriptsize{$\pm$ 7.0} & 25.4 \scriptsize{$\pm$ 1.7} & 89.9 \scriptsize{$\pm$ 2.7} & 62.0 \scriptsize{$\pm$ 5.2} & 54.1 \scriptsize{$\pm$ 2.8} & 33.1 \scriptsize{$\pm$ 2.1} & 50.0 \scriptsize{$\pm$ 5.0} & 52.5 \scriptsize{$\pm$ 1.9} & 3.8 \scriptsize{$\pm$ 0.4} \cr
\rowcolor{Gray} RetinaNet & \cellcolor{White} &12.6 \scriptsize{$\pm$ 0.6} & 69.2 \scriptsize{$\pm$ 2.5} & 64.3 \scriptsize{$\pm$ 4.5} & 40.9 \scriptsize{$\pm$ 1.6} & 21.5 \scriptsize{$\pm$ 0.5} & 90.2 \scriptsize{$\pm$ 1.7} & 60.4 \scriptsize{$\pm$ 3.5} & 50.0 \scriptsize{$\pm$ 1.5} & 32.7 \scriptsize{$\pm$ 1.4} & 46.3 \scriptsize{$\pm$ 2.4} & 48.8 \scriptsize{$\pm$ 1.2} & 5.2 \scriptsize{$\pm$ 0.5} \cr
Deformable-DETR& \cellcolor{White} & 18.3 \scriptsize{$\pm$ 0.9} & 78.6 \scriptsize{$\pm$ 2.0} & 70.3 \scriptsize{$\pm$ 3.4} & 54.5 \scriptsize{$\pm$ 1.5} & 24.2 \scriptsize{$\pm$ 1.1} & 86.7 \scriptsize{$\pm$ 2.2} & 61.1 \scriptsize{$\pm$ 3.9} & 59.6 \scriptsize{$\pm$ 1.7} & 39.8 \scriptsize{$\pm$ 3.3} & 54.3 \scriptsize{$\pm$ 2.8} & 54.7 \scriptsize{$\pm$ 1.0} & 2.7 \scriptsize{$\pm$ 0.5} \cr
 \rowcolor{Gray}  Cascade R-CNN-P67 & \cellcolor{White} \multirowcell{-6}{COCO} & 19.1 \scriptsize{$\pm$ 0.8} & 73.3 \scriptsize{$\pm$ 1.6} & 75.4 \scriptsize{$\pm$ 2.5} & 63.4 \scriptsize{$\pm$ 1.0} & 27.4 \scriptsize{$\pm$ 1.2} & 91.2 \scriptsize{$\pm$ 2.1} & 65.7 \scriptsize{$\pm$ 4.7} & 56.2 \scriptsize{$\pm$ 1.6} & 38.3 \scriptsize{$\pm$ 2.4} & 53.7 \scriptsize{$\pm$ 2.9} & \textbf{56.4} \scriptsize{$\pm$ 1.1} & \textbf{1.7} \scriptsize{$\pm$ 0.4} \cr
\cmidrule(lr){1-1}\cmidrule(lr){2-2}\cmidrule(lr){3-14}
Faster R-CNN & \cellcolor{Gray} &  17.7 \scriptsize{$\pm$ 0.7} & 74.8 \scriptsize{$\pm$ 1.9} & 71.9 \scriptsize{$\pm$ 2.5} & 51.1 \scriptsize{$\pm$ 1.0} & 31.0 \scriptsize{$\pm$ 0.8} & 90.5 \scriptsize{$\pm$ 1.6} & 63.1 \scriptsize{$\pm$ 4.2} & 51.1 \scriptsize{$\pm$ 1.7} & 36.6 \scriptsize{$\pm$ 2.0} & 47.7 \scriptsize{$\pm$ 2.6} & 53.6 \scriptsize{$\pm$ 1.0} & 2.1 \scriptsize{$\pm$ 0.3} \cr
\rowcolor{Gray} CenterNet2 & \cellcolor{Gray} & 16.6 \scriptsize{$\pm$ 0.8} & 74.6 \scriptsize{$\pm$ 2.0} & 68.1 \scriptsize{$\pm$ 3.1} & 43.9 \scriptsize{$\pm$ 1.3} & 28.3 \scriptsize{$\pm$ 0.9} & 90.8 \scriptsize{$\pm$ 2.0} & 64.1 \scriptsize{$\pm$ 4.1} & 50.0 \scriptsize{$\pm$ 1.6} & 34.1 \scriptsize{$\pm$ 1.7} & 45.7 \scriptsize{$\pm$ 2.9} & 51.6 \scriptsize{$\pm$ 1.0} & 2.7 \scriptsize{$\pm$ 0.3} \cr
Cascade R-CNN-P67 & \cellcolor{Gray} \multirow{-3}{*}{LVIS} & 18.6 \scriptsize{$\pm$ 0.7} & 76.1 \scriptsize{$\pm$ 1.4} & 72.8 \scriptsize{$\pm$ 2.7} & 55.7 \scriptsize{$\pm$ 1.1} & 33.1 \scriptsize{$\pm$ 0.6} & 91.4 \scriptsize{$\pm$ 2.0} & 66.9 \scriptsize{$\pm$ 4.0} & 52.5 \scriptsize{$\pm$ 1.5} & 37.7 \scriptsize{$\pm$ 2.6} & 51.1 \scriptsize{$\pm$ 2.8} & \textbf{55.6} \scriptsize{$\pm$ 1.0} & \textbf{1.2} \scriptsize{$\pm$ 0.3} \cr
\bottomrule
\end{tabular}
}
\caption{Performance of different architectures pre-trained on COCO and LVIS.}
\label{tab:arch_10shot}
\end{subtable}

\begin{subtable}[h]{\textwidth}
\resizebox{\linewidth}{!}{
\begin{tabular}{c@{\hskip 6pt}ccccccccccccc}
\toprule
\multicolumn{2}{c}{10-shot} & Aerial & Agriculture & Animal & Cartoon & Fashion & Food & Logo & Person & Security & Traffic & \multirowcell{2}{Mean} & \multirowcell{2}{Rank} \cr \smidrulea
Architecture & Pre-training & VisDrone & DeepFruits & iWildCam & Clipart & iMaterialist & Oktoberfest & LogoDet-3K & CrowdHuman & SIXray & KITTI \cr \smidruleb
& ImageNet &13.5 \scriptsize{$\pm$ 0.5} & 66.5 \scriptsize{$\pm$ 2.5} & 14.6 \scriptsize{$\pm$ 4.4} & 25.8 \scriptsize{$\pm$ 1.6} & 21.9 \scriptsize{$\pm$ 0.6} & 86.2 \scriptsize{$\pm$ 3.1} & 54.7 \scriptsize{$\pm$ 3.7} & 38.9 \scriptsize{$\pm$ 1.3} & 22.1 \scriptsize{$\pm$ 2.3} & 32.6 \scriptsize{$\pm$ 3.0} & 37.7 \scriptsize{$\pm$ 1.2} & 5.9 \scriptsize{$\pm$ 0.2} \cr
\rowcolor{Gray} \cellcolor{White} & COCO & 19.1 \scriptsize{$\pm$ 0.8} & 73.3 \scriptsize{$\pm$ 1.6} & 75.4 \scriptsize{$\pm$ 2.5} & 63.4 \scriptsize{$\pm$ 1.0} & 27.4 \scriptsize{$\pm$ 1.2} & 91.2 \scriptsize{$\pm$ 2.1} & 65.7 \scriptsize{$\pm$ 4.7} & 56.2 \scriptsize{$\pm$ 1.6} & 38.3 \scriptsize{$\pm$ 2.4} & 53.7 \scriptsize{$\pm$ 2.9} & 56.4 \scriptsize{$\pm$ 1.1} & 2.9 \scriptsize{$\pm$ 0.4} \cr
\cellcolor{White} &FSODD & 13.7 \scriptsize{$\pm$ 0.6} & 74.9 \scriptsize{$\pm$ 2.1} & 67.8 \scriptsize{$\pm$ 3.7} & 35.9 \scriptsize{$\pm$ 1.9} & 24.6 \scriptsize{$\pm$ 0.9} & 90.8 \scriptsize{$\pm$ 1.5} & 66.5 \scriptsize{$\pm$ 3.6} & 42.9 \scriptsize{$\pm$ 1.8} & 34.3 \scriptsize{$\pm$ 1.6} & 39.5 \scriptsize{$\pm$ 2.6} & 49.1 \scriptsize{$\pm$ 1.0} & 4.5 \scriptsize{$\pm$ 0.4} \cr
\rowcolor{Gray} \cellcolor{White} & LVIS & 18.6 \scriptsize{$\pm$ 0.7} & 76.1 \scriptsize{$\pm$ 1.4} & 72.8 \scriptsize{$\pm$ 2.7} & 55.7 \scriptsize{$\pm$ 1.1} & 33.1 \scriptsize{$\pm$ 0.6} & 91.4 \scriptsize{$\pm$ 2.0} & 66.9 \scriptsize{$\pm$ 4.0} & 52.5 \scriptsize{$\pm$ 1.5} & 37.7 \scriptsize{$\pm$ 2.6} & 51.1 \scriptsize{$\pm$ 2.8} & 55.6 \scriptsize{$\pm$ 1.0} & 3.2 \scriptsize{$\pm$ 0.5} \cr
\cellcolor{White} & Unified & 19.7 \scriptsize{$\pm$ 0.7} & 76.8 \scriptsize{$\pm$ 1.1} & 69.9 \scriptsize{$\pm$ 3.4} & 58.6 \scriptsize{$\pm$ 1.5} & 29.5 \scriptsize{$\pm$ 1.1} & 92.8 \scriptsize{$\pm$ 1.0} & 69.5 \scriptsize{$\pm$ 4.0} & 55.6 \scriptsize{$\pm$ 1.5} & 39.9 \scriptsize{$\pm$ 2.9} & 53.6 \scriptsize{$\pm$ 2.9} & 56.6 \scriptsize{$\pm$ 1.1} & 2.4 \scriptsize{$\pm$ 0.2} \cr
\rowcolor{Gray} \cellcolor{White} \multirow{-6}{*}{Cascade R-CNN-P67~} & LVIS+ & 21.4 \scriptsize{$\pm$ 0.7} & 84.4 \scriptsize{$\pm$ 1.5} & 67.1 \scriptsize{$\pm$ 3.5} & 60.4 \scriptsize{$\pm$ 0.8} & 34.4 \scriptsize{$\pm$ 0.5} & 89.9 \scriptsize{$\pm$ 1.8} & 70.7 \scriptsize{$\pm$ 3.2} & 55.1 \scriptsize{$\pm$ 1.4} & 50.2 \scriptsize{$\pm$ 2.6} & 55.9 \scriptsize{$\pm$ 2.9} & \textbf{59.0} \scriptsize{$\pm$ 1.0} & \textbf{2.0} \scriptsize{$\pm$ 0.4} \cr
\cmidrule(lr){1-1}\cmidrule(lr){2-2}\cmidrule(lr){3-14}
\cellcolor{Gray} & COCO & 7.9 \scriptsize{$\pm$ 0.8} & 35.6 \scriptsize{$\pm$ 5.2} & 38.2 \scriptsize{$\pm$ 20.6} & 33.0 \scriptsize{$\pm$ 7.7} & 14.7 \scriptsize{$\pm$ 2.2} & 45.2 \scriptsize{$\pm$ 6.7} & 27.6 \scriptsize{$\pm$ 4.3} & 43.8 \scriptsize{$\pm$ 4.8} & 7.0 \scriptsize{$\pm$ 4.1} & 38.4 \scriptsize{$\pm$ 3.6} & 29.1 \scriptsize{$\pm$ 5.2} & 3.1 \scriptsize{$\pm$ 0.4} \cr
\rowcolor{Gray} & LVIS & 7.6 \scriptsize{$\pm$ 0.7} & 41.5 \scriptsize{$\pm$ 6.0} & 36.0 \scriptsize{$\pm$ 12.7} & 20.0 \scriptsize{$\pm$ 2.4} & 18.7 \scriptsize{$\pm$ 1.4} & 50.4 \scriptsize{$\pm$ 7.1} & 27.8 \scriptsize{$\pm$ 5.0} & 38.7 \scriptsize{$\pm$ 2.6} & 8.6 \scriptsize{$\pm$ 4.2} & 32.3 \scriptsize{$\pm$ 3.6} & 28.1 \scriptsize{$\pm$ 3.3} & 3.3 \scriptsize{$\pm$ 0.3} \cr
\cellcolor{Gray} & LVIS+ & 10.6 \scriptsize{$\pm$ 1.1} & 55.7 \scriptsize{$\pm$ 5.4} & 38.4 \scriptsize{$\pm$ 12.8} & 35.5 \scriptsize{$\pm$ 2.9} & 25.1 \scriptsize{$\pm$ 1.1} & 48.4 \scriptsize{$\pm$ 6.7} & 36.7 \scriptsize{$\pm$ 4.2} & 46.4 \scriptsize{$\pm$ 3.3} & 13.9 \scriptsize{$\pm$ 5.7} & 37.9 \scriptsize{$\pm$ 5.3} & 34.9 \scriptsize{$\pm$ 3.2} & 1.9 \scriptsize{$\pm$ 0.3} \cr
\rowcolor{Gray} \multirow{-4}{*}{CenterNet2} & LVIS++ & 10.7 \scriptsize{$\pm$ 1.0} & 59.4 \scriptsize{$\pm$ 5.5} & 41.7 \scriptsize{$\pm$ 13.4} & 38.2 \scriptsize{$\pm$ 2.3} & 26.7 \scriptsize{$\pm$ 1.0} & 46.2 \scriptsize{$\pm$ 6.1} & 35.6 \scriptsize{$\pm$ 4.2} & 47.0 \scriptsize{$\pm$ 3.5} & 15.7 \scriptsize{$\pm$ 5.8} & 37.1 \scriptsize{$\pm$ 4.7} & \textbf{35.8} \scriptsize{$\pm$ 3.4} & \textbf{1.7} \scriptsize{$\pm$ 0.3} \cr
\bottomrule
\end{tabular}
}
\caption{Performance of Cascade R-CNN-P67 and CenterNet2 pre-trained on different datasets.}
\label{tab:pre_training_10shot}
\end{subtable}

\end{table*}

%% file: table_dataset.tex
\begin{table*}[h]
\renewcommand{\thempfootnote}{\fnsymbol{mpfootnote}}
\def\smidrule{\cmidrule(rl){1-1}\cmidrule(rl){2-2}\cmidrule(rl){3-3}\cmidrule(rl){4-4}\cmidrule(rl){5-5}\cmidrule(rl){6-6}\cmidrule(rl){7-7}}
\caption{Statistics of 32 datasets in the extended benchmark. The 10 datasets used in MoFSOD are shown in \textbf{bold}.}
\label{tb:dataset}
\begin{minipage}[h]{\linewidth}
\centering
\resizebox{0.85\linewidth}{!}{%
\begin{tabular}{ccccccc}
\toprule
Domain & Dataset & \# classes & \# train images & \# train anno.  & \# test images & \# test anno. \cr \smidrule
\multirowcell{3}{Aerial} & CARPK & 1 & 989 & 42275 & 459 & 47501 \cr
 & DOTA & 15 & 8949 & 116515 & 8949 & 116515 \cr
 & \textbf{VisDrone} & 10 & 7019 & 381965 & 1610 & 75103 \cr \smidrule
\multirowcell{2}{Agriculture} & \textbf{DeepFruits} & 7 & 457 & 2553 & 114 & 590 \cr
 & MinneApple & 1 & 403 & 19373 & 267 & 8811 \cr \smidrule
\multirowcell{2}{Animal} & ENA24 & 22 & 7031 & 7811 & 1758 & 1963 \cr
 & \textbf{iWildCam} & 1 & 21065 & 31591 & 5313 & 7901 \cr \smidrule
\multirowcell{3}{Cartoon} & \textbf{Clipart} & 20 & 500 & 1640 & 500 & 1527 \cr
 & Comic & 6 & 1000 & 3215 & 1000 & 3176 \cr
 & Watercolor & 6 & 1000 & 1662 & 1000 & 1655 \cr \smidrule
Dense Product & SKU110K & 1 & 8804 & 1298968 & 2935 & 431420 \cr \smidrule
Face & WIDER FACE & 1 & 12880 & 159423 & 3222 & 39698 \cr \smidrule
\multirowcell{2}{Fashion} & DeepFashion2 & 13 & 191961 & 312187 & 32153 & 52491 \cr
 & \textbf{iMaterialist} & 46 & 45623 & 333402 & 1158 & 8782 \cr \smidrule
\multirowcell{2}{Food} & Kitchen & 11 & 4711 & 24730 & 2016 & 13430 \cr
 & \textbf{Oktoberfest} & 15 & 1110 & 2697 & 85 & 236 \cr \smidrule
Head & HollywoodHeads & 1 & 10834 & 17754 & 3984 & 7080 \cr \smidrule
\multirowcell{2}{Logo} & \textbf{LogoDet-3K} & 2993 & 126891 & 155286 & 31727 & 38981 \cr
 & OpenLogo & 352 & 18752 & 35264 & 8331 & 15945 \cr \smidrule
\multirowcell{2}{Medical} & ChestX-Det10 & 10 & 2320 & 6864 & 459 & 1477 \cr
 & DeepLesion & 1 & 27289 & 28871 & 4831 & 5122 \cr \smidrule
\multirowcell{2}{Person} & \textbf{CrowdHuman} & 2 & 15000 & 705967 & 4370 & 206231 \cr
 & WiderPerson & 1 & 8000 & 245053 & 1000 & 28424 \cr \smidrule
\multirowcell{2}{Security} & PIDray & 12 & 29454 & 39709 & 9482 & 9483 \cr
 & \textbf{SIXray} & 5 & 7496 & 15439 & 1310 & 2054 \cr \smidrule
Table & table-detection & 1 & 212 & 244 & 191 & 276 \cr \smidrule
Text & COCO-Text & 2 & 19039 & 163477 & 4446 & 37651 \cr \smidrule
\multirowcell{4}{Traffic} & Cityscapes & 8 & 2965 & 50348 & 492 & 9793 \cr
 & \textbf{KITTI} & 4 & 5481 & 38077 & 7481 & 52458 \cr
 & LISA & 5 & 7937 & 9246 & 1987 & 2283 \cr
 & TT100K & 151 & 6105 & 16528 & 3071 & 8175 \cr \smidrule
Underwater & DUO & 4 & 6617 & 63999 & 1100 & 10518 \cr
\bottomrule
\end{tabular}
}
\end{minipage}
\end{table*}

%% file: table_all.tex
\begin{table*}[t]
\def\smidrule{\cmidrule(rl){1-1}\cmidrule(rl){2-2}\cmidrule(rl){3-3}\cmidrule(rl){4-4}\cmidrule(rl){5-5}\cmidrule(rl){6-6}}
\centering
\caption{Performance on the proposed benchmark in AP50 (top) and the average rank (bottom) of different methods on the extended benchmark with 32 datasets. Note that the pre-trained model used for this table is different from the main paper, such that the comparison is fair within these tables only.}
\label{tab:result_all}
%
\scalebox{0.75}{%
\begin{tabular}{cccccc}
\toprule
Method & 1-shot & 3-shot & 5-shot & 10-shot & Mean \cr \smidrule
TFA~\cite{wang2020frustratingly} &
20.6 \scriptsize{$\pm$ 3.7} & 25.6 \scriptsize{$\pm$ 2.0} & 27.9 \scriptsize{$\pm$ 1.7} & 30.9 \scriptsize{$\pm$ 1.3} & 26.3 \scriptsize{$\pm$ 2.4} \cr
\rowcolor{Gray} TFA-cos~\cite{wang2020frustratingly} &
20.8 \scriptsize{$\pm$ 3.6} & 25.6 \scriptsize{$\pm$ 2.0} & 27.7 \scriptsize{$\pm$ 1.7} & 30.5 \scriptsize{$\pm$ 1.3} & 26.1 \scriptsize{$\pm$ 2.4} \cr
FSCE-base~\cite{sun2021fsce} &
25.3 \scriptsize{$\pm$ 4.2} & 33.6 \scriptsize{$\pm$ 3.9} & 37.9 \scriptsize{$\pm$ 2.0} & 43.4 \scriptsize{$\pm$ 1.6} & 35.1 \scriptsize{$\pm$ 3.3} \cr
\rowcolor{Gray} FSCE-con~\cite{sun2021fsce} &
25.8 \scriptsize{$\pm$ 4.4} & 34.1 \scriptsize{$\pm$ 3.7} & 38.1 \scriptsize{$\pm$ 1.8} & 43.3 \scriptsize{$\pm$ 1.5} & 35.3 \scriptsize{$\pm$ 3.3} \cr
DeFRCN~\cite{qiao2021defrcn} &
25.4 \scriptsize{$\pm$ 4.0} & 32.9 \scriptsize{$\pm$ 3.1} & 36.7 \scriptsize{$\pm$ 1.9} & 41.2 \scriptsize{$\pm$ 1.9} & 34.1 \scriptsize{$\pm$ 3.0} \cr
\rowcolor{Gray} FT &
\textbf{26.2} \scriptsize{$\pm$ 3.3} & \textbf{35.2} \scriptsize{$\pm$ 3.5} & \textbf{39.6} \scriptsize{$\pm$ 2.3} & \textbf{45.8} \scriptsize{$\pm$ 2.1} & \textbf{36.7} \scriptsize{$\pm$ 2.9} \cr
\bottomrule
\end{tabular}
}
%
\scalebox{0.75}{%
\begin{tabular}{ccccccc}
\toprule
Method & 1-shot & 3-shot & 5-shot & 10-shot & Mean \cr \smidrule
TFA~\cite{wang2020frustratingly} &
4.7 \scriptsize{$\pm$ 0.5} & 4.8 \scriptsize{$\pm$ 0.4} & 4.7 \scriptsize{$\pm$ 0.3} & 4.7 \scriptsize{$\pm$ 0.4} & 4.7 \scriptsize{$\pm$ 0.4} \cr
\rowcolor{Gray} TFA-cos~\cite{wang2020frustratingly} &
4.3 \scriptsize{$\pm$ 0.4} & 4.5 \scriptsize{$\pm$ 0.4} & 4.7 \scriptsize{$\pm$ 0.4} & 4.8 \scriptsize{$\pm$ 0.4} & 4.6 \scriptsize{$\pm$ 0.4} \cr
FSCE-base~\cite{sun2021fsce} &
3.3 \scriptsize{$\pm$ 0.4} & 3.0 \scriptsize{$\pm$ 0.4} & 2.9 \scriptsize{$\pm$ 0.3} & 2.8 \scriptsize{$\pm$ 0.3} & 3.0 \scriptsize{$\pm$ 0.4} \cr
\rowcolor{Gray} FSCE-con~\cite{sun2021fsce} &
2.8 \scriptsize{$\pm$ 0.4} & 2.7 \scriptsize{$\pm$ 0.3} & 2.7 \scriptsize{$\pm$ 0.3} & 2.7 \scriptsize{$\pm$ 0.3} & 2.7 \scriptsize{$\pm$ 0.4} \cr
DeFRCN~\cite{qiao2021defrcn} &
3.3 \scriptsize{$\pm$ 0.5} & 3.5 \scriptsize{$\pm$ 0.5} & 3.5 \scriptsize{$\pm$ 0.4} & 3.7 \scriptsize{$\pm$ 0.3} & 3.5 \scriptsize{$\pm$ 0.5} \cr
\rowcolor{Gray} FT &
\textbf{2.7} \scriptsize{$\pm$ 0.4} & \textbf{2.5} \scriptsize{$\pm$ 0.4} & \textbf{2.4} \scriptsize{$\pm$ 0.5} & \textbf{2.2} \scriptsize{$\pm$ 0.5} & \textbf{2.5} \scriptsize{$\pm$ 0.5} \cr
\bottomrule
\end{tabular}
}
%
\end{table*}

%% file: table_arch.tex
\begin{table*}[t]
\def\smidrule{\cmidrule(rl){1-1}\cmidrule(rl){2-2}\cmidrule(rl){3-3}\cmidrule(rl){4-4}\cmidrule(rl){5-5}\cmidrule(rl){6-6}\cmidrule(rl){7-7}}
\centering
\caption{Performance of FT on the proposed benchmark with different model architectures and pre-training datasets in AP50 (top) and the average rank (bottom) on the 32 datasets extended benchmark. Note that the pre-trained model used for this table is different from the main paper, such that the comparison is fair within these tables only.}
\label{tab:result_arch}
%
\scalebox{0.75}{%
\begin{tabular}{ccccccc}
\toprule
Arch & Pre-train & 1-shot & 3-shot & 5-shot & 10-shot & Mean \cr \smidrule
Faster R-CNN & COCO &
26.2 \scriptsize{$\pm$ 3.3} & 35.2 \scriptsize{$\pm$ 3.5} & 39.6 \scriptsize{$\pm$ 2.3} & 45.8 \scriptsize{$\pm$ 2.1} & 36.7 \scriptsize{$\pm$ 2.9} \cr
\rowcolor{Gray} & COCO &
26.4 \scriptsize{$\pm$ 3.3} & 35.7 \scriptsize{$\pm$ 2.6} & 40.3 \scriptsize{$\pm$ 2.0} & 46.7 \scriptsize{$\pm$ 1.6} & 37.3 \scriptsize{$\pm$ 2.5} \cr
\rowcolor{White} \cellcolor{Gray} & FSODD &
23.3 \scriptsize{$\pm$ 3.2} & 32.4 \scriptsize{$\pm$ 2.5} & 36.8 \scriptsize{$\pm$ 1.8} & 42.9 \scriptsize{$\pm$ 1.7} & 33.8 \scriptsize{$\pm$ 2.5} \cr
\rowcolor{Gray} \multirow{-3}{*}{Cascade R-CNN-P67~} & Unified &
\textbf{29.2} \scriptsize{$\pm$ 2.9} & \textbf{38.7} \scriptsize{$\pm$ 2.6} & \textbf{43.4} \scriptsize{$\pm$ 1.6} & \textbf{49.7} \scriptsize{$\pm$ 1.9} & \textbf{40.3} \scriptsize{$\pm$ 2.4} \cr
\bottomrule
\end{tabular}
}
%
%
\scalebox{0.75}{%
\begin{tabular}{cccccccc}
\toprule
Arch & Pre-train & 1-shot & 3-shot & 5-shot & 10-shot & Mean \cr \smidrule
Faster R-CNN & COCO &
2.8 \scriptsize{$\pm$ 0.4} & 2.9 \scriptsize{$\pm$ 0.4} & 3.0 \scriptsize{$\pm$ 0.3} & 3.0 \scriptsize{$\pm$ 0.3} & 2.9 \scriptsize{$\pm$ 0.3} \cr
\rowcolor{Gray} & COCO &
2.5 \scriptsize{$\pm$ 0.3} & 2.5 \scriptsize{$\pm$ 0.2} & 2.5 \scriptsize{$\pm$ 0.3} & 2.4 \scriptsize{$\pm$ 0.3} & 2.5 \scriptsize{$\pm$ 0.3} \cr
\rowcolor{White} \cellcolor{Gray} & FSODD &
3.0 \scriptsize{$\pm$ 0.3} & 3.1 \scriptsize{$\pm$ 0.3} & 3.3 \scriptsize{$\pm$ 0.3} & 3.4 \scriptsize{$\pm$ 0.3} & 3.2 \scriptsize{$\pm$ 0.3} \cr
\rowcolor{Gray} \multirow{-3}{*}{Cascade R-CNN-P67~} & Unified &
\textbf{1.6} \scriptsize{$\pm$ 0.4} & \textbf{1.4} \scriptsize{$\pm$ 0.3} & \textbf{1.3} \scriptsize{$\pm$ 0.3} & \textbf{1.2} \scriptsize{$\pm$ 0.3} & \textbf{1.4} \scriptsize{$\pm$ 0.3} \cr
\bottomrule
\end{tabular}
}
%
\end{table*}